\documentclass[sigconf,nonacm]{acmart}

\usepackage{multirow}
\usepackage{subfigure}
\usepackage{color}
\usepackage[utf8]{inputenc}
\usepackage{longtable}
\usepackage{lscape}

\begin{document}

\title{A Large-Scale Benchmark for Food Image Segmentation}

\author{Xiongwei Wu}
\affiliation{%
  \institution{Singapore Management University}
  \country{}
}
\email{xwwu@smu.edu.sg}

\author{Xin Fu}
\affiliation{%
  \institution{Beijing Jiaotong University}
  \country{}
}
\email{xinfu@bjtu.edu.cn}

\author{Ying Liu}
\affiliation{%
  \institution{Singapore Management University}
  \country{}
}
\email{rrrainbowly@gmail.com}

\author{Ee-Peng Lim}
\affiliation{%
  \institution{Singapore Management University}
  \country{}
}
\email{eplim@smu.edu.sg}

\author{Steven C.H. Hoi}
\affiliation{%
  \institution{Salesforce Research Asia }
  \institution{Singapore Management University}
  \country{}
}
\email{chhoi@smu.edu.sg}

\author{Qianru Sun}
\affiliation{%
  \institution{Singapore Management University}
  \country{}
}
\email{qianrusun@smu.edu.sg}

\begin{abstract}
Food image segmentation is a critical and indispensible task for developing health-related applications such as estimating food calories and nutrients. Existing food image segmentation models are underperforming due to two reasons: (1) there is a lack of high quality food image datasets with fine-grained ingredient labels and pixel-wise location masks---the existing datasets either carry coarse ingredient labels or are small in size; and (2) the complex appearance of food makes it difficult to localize and recognize ingredients in food images, e.g., the ingredients may overlap one another in the same image, and the identical ingredient may appear distinctly in different food images. 

In this work, we build a new food image dataset FoodSeg103 (and its extension FoodSeg154) containing 9,490 images. We annotate these images with 154 ingredient classes and each image has an average of 6 ingredient labels and pixel-wise masks.
In addition, we propose a multi-modality pre-training approach called ReLeM that explicitly equips a segmentation model with rich and semantic food knowledge.
In experiments, we use three popular semantic segmentation methods (i.e., Dilated Convolution based~\cite{huang2018ccnet}, Feature Pyramid based~\cite{kirillov2019panoptic}, and Vision Transformer based~\cite{SETR}) as baselines, and evaluate them as well as ReLeM on our new datasets.
We believe that the FoodSeg103 (and its extension FoodSeg154) and the pre-trained models using ReLeM can serve as a benchmark to facilitate future works on fine-grained food image understanding. We make all these datasets and methods public at \url{https://xiongweiwu.github.io/foodseg103.html}.

\end{abstract}

\maketitle

\section{Introduction}
\begin{figure}[t]
\vspace{3mm}
	\centering{
	\subfigure{
		\hspace{-1.0mm}\includegraphics[width=0.48\textwidth]{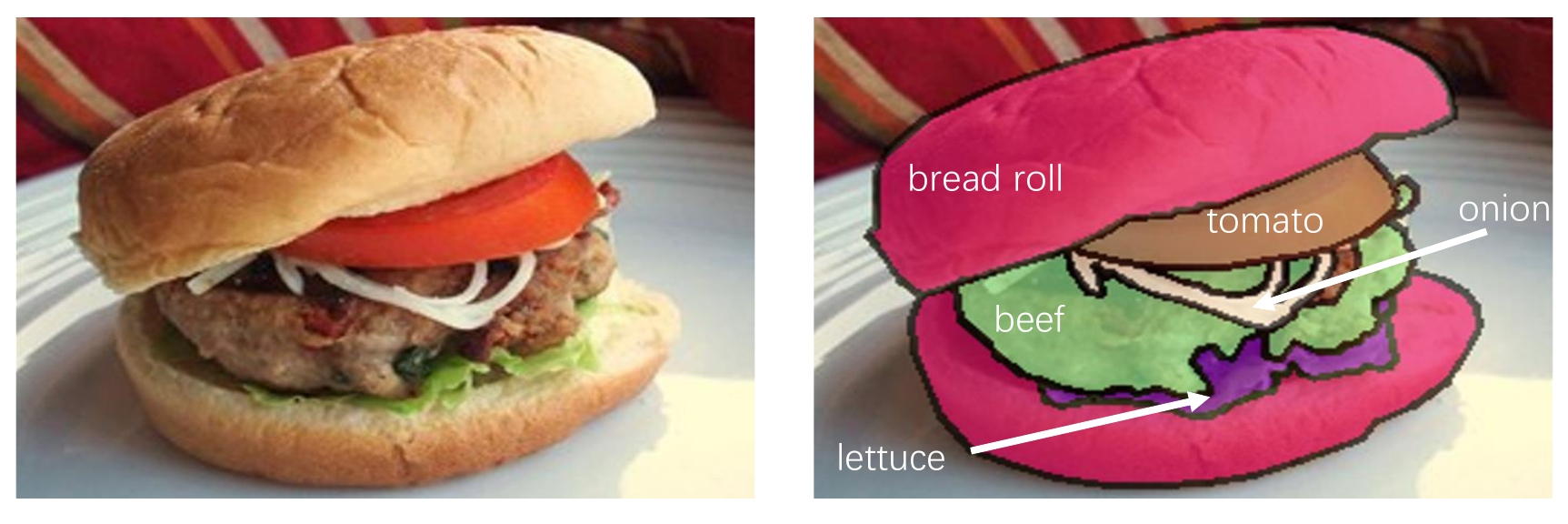}}\vspace{-3mm}
	\subfigure{
		\includegraphics[width=0.15\textwidth]{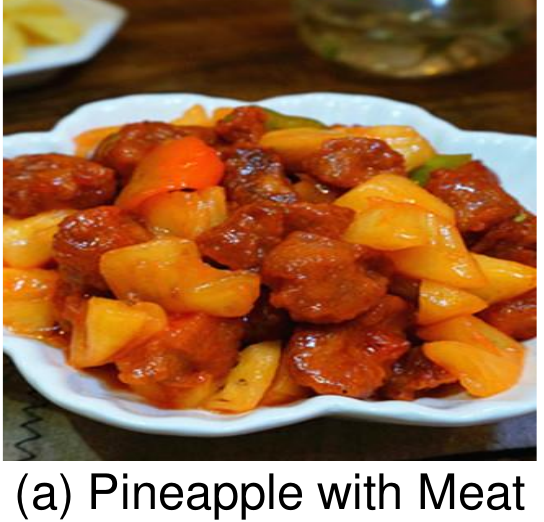}}
	\subfigure{
		\includegraphics[width=0.15\textwidth]{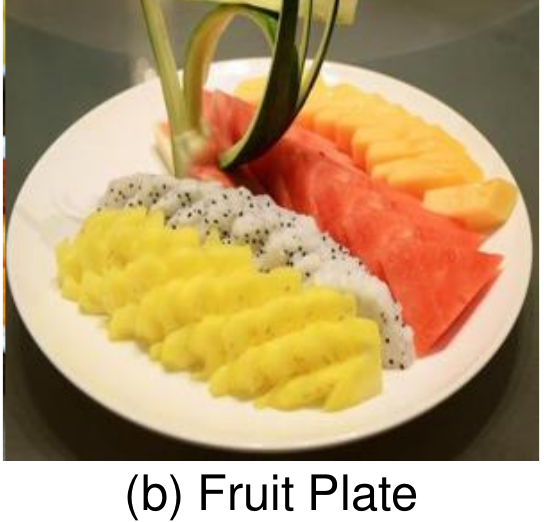}}
	\subfigure{
		\includegraphics[width=0.15\textwidth]{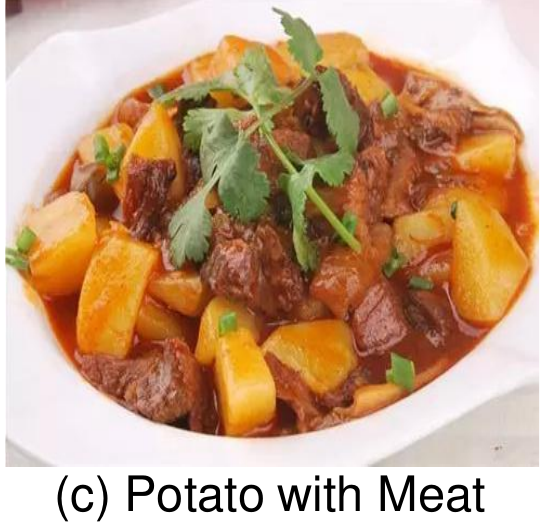}}
		}
	\caption{The first row shows a source image and its segmentation masks on our FoodSeg103.
	The second row shows example images to reveal the difficulties of food image segmentation, e.g., the pineapples in (a) and (b) look different, while the pineapple in (a) and the potato in (c) look quite similar.} 	\label{fig:example}
	\vspace{-4mm}
\end{figure}

Food computing has attracted increasing public attention in recent years, as it provides the core technologies for food and health-related research and applications.~\cite{David-DH-Nature2014,Boswell-FC-PNAS2018,Meyers-Im2Calories-ICCV2015,Quin-Nutrition5k-CVPR2021}. 
One of the important goals of food computing is to automatically recognize different types of food and profile their nutrition and calorie values.
In computer vision, the related works include 
dish classification~\cite{deng2019mixed,xu2015geolocalized,wang2019mixed}, recipe generation~\cite{salvador2019inverse,wang2020structure,h2020recipegpt}, and food image retrieval~\cite{shimoda2017learning,ciocca2017learning}. 
Most of them focus on representing and analysing the food image as a whole, and do not explicitly localize or classify its individual ingredients---the visible components in the cooked food. We call the former food image classification and the latter food image segmentation.
Between the two, food image segmentation is more complex as it aims to recognize each ingredient category as well as its pixel-wise locations in the food image.
As shown in Figure~\ref{fig:example}, given an ``hamburger'' example image, a good segmentation model needs to recognize and mask out  ``beef'', ``tomato'', ``lettuce'', ``onion'' and ``bread roll'' ingredients.

Compared to semantic segmentation on general object images~\cite{chen2014semantic,huang2018ccnet,kirillov2019panoptic}, food image segmentation is more challenging due to the large diversity in food appearances and the often imbalanced distribution of categories of ingredients. First, an ingredient cooked differently can vary a lot visually, e.g., ``pineapples'' cooked with meat in Figure~\ref{fig:example} (a) versus the ``pineapples'' in a fruit platter in Figure~\ref{fig:example} (b). Different ingredients may look very similar, e.g., ``pineapples'' cooked with meat cannot be easily distinguished from ``potatoes'' cooked with meat, as shown in Figures~\ref{fig:example} (a) and (c) respectively.
Second, food datasets usually suffer from imbalanced distribution---both food classes and ingredient classes often exist in long-tailed distributions. This is inevitable due to two reasons: 1) large number of food images are dominated by very few popular food classes while vast majority of food classes are unpopular; and 2) there is a selection bias in the construction of food image collection~\cite{torralba2011unbiased}.
We will elaborate the detailed distribution analysis in Section~\ref{sec:fn}.

Existing food image datasets, such as ETH Food101 \cite{Bossard-Food101-ECCV2014}, Recipe1M \cite{salvador2017learning}, and Geo-Dish \cite{xu2015geolocalized}, mainly facilitate the research of dish classification or recipe generation.
They do not have fine-grained ingredient masks or labels.
UECFoodPix \cite{uecfoodpix} and UECFoodPixComplete \cite{uecfoodpixcomplete} are the only two public datasets for food image segmentation. However, their segmentation masks are annotated at dish level only. That is, each mask covers the region of an entire dish instead of that of food ingredients. We elaborate more dataset comparison in Section~\ref{sec:comp}.

\textbf{Dataset contribution:}
To facilitate fine-grained food image segmentation, we build a large-scale dataset called FoodSeg103, for which we have defined 103 ingredient classes and annotated 7,118 western food images using these labels together with the corresponding segmentation masks.
Besides, we annotated an additional set of 2,372 images of Asian food which covers more diverse set of ingredients making these images more challenging than those in the main set (FoodSeg103). For this set, we defined 112 ingredient classes---55\% overlap with the ingredient classes of the main set. 
In total, we annotated 154 classes of ingredients with around 60k masks (in the two datasets). We name the combined dataset as FoodSeg154. 
During the annotation, we carried out careful data selection, iterative refinement of labels and masks (to be further elaborated in Section~\ref{sec:collect}), so as to guarantee high quality labels and masks in the dataset.
Our annotation is thus expensive and time-consuming. 
In experiments, we use FoodSeg103 for in-domain training and testing, and use the additional set in FoodSeg154 for out-domain testing.

\textbf{Model contribution:}
The source images of FoodSeg103 are from another existing food dataset Recipe1M~\cite{salvador2017learning}---millions of images and cooking recipes, used for recipe generation. Each recipe contains not only ``how to cook'' but also ``what ingredient to use''.
In our work, we leverage these recipe information as auxiliary information to train semantic segmentation models.
We call this \textit{multi-modality knowledge transfer} and name our training method ReLeM.
Specifically, ReLeM 
integrates food recipe data, in the format of language embedding, with the visual representation of the food image.
In this way, it forces the visual representation of an ingredient appearing in different dishes to have their appearances ``connected'' in the feature space through a common language embedding (extracted from the ingredient's label and its cooking instructions).

\textbf{Experiment contribution:} We validate our proposed ReLeM model by plugging it into the  state-of-the-art semantic segmentation models such as CCNet~\cite{huang2018ccnet}, Sem-FPN~\cite{kirillov2019panoptic} and SeTR~\cite{SETR}. In experiments, we compare ReLeM-variants with these baseline models using both convolutional networks and transformer backbones. Our experiments show that ReLeM is generic to be applied into multiple segmentation frameworks, and it helps to achieve significant accuracy improvement when incorporated into the SOTA CNN-based model CCNet. This validates that our knowledge transfer approach works more efficient on stronger models---a characteristic preferred by the multimedia community.

Our contributions are thus three-fold.
i) We build a large-scale food image segmentation dataset called FoodSeg103 (and its extension FoodSeg154). It can facilitate a promising and challenging benchmark for the task of semantic segmentation in food images.
ii) We propose a knowledge transfer approach ReLeM that utilizes the multi-modality information of recipe datasets. 
It can be incorporated into different semantic segmentation methods to boost the model performance.
iii) We conduct extensive experiments that reveal the challenges of segmenting food on our FoodSeg103 dataset, and validate the efficiency of our ReLeM based on multiple baseline methods.

\section{Related Works}~\label{sec:rw}

\noindent{\textbf{Food Image Datasets.}} 
In recent years, the scale of food-related datasets has grown rapidly. 
For example, Bossard et al~\cite{Bossard-Food101-ECCV2014} built one large-scale food dataset ETH Food101, which contains 101 classes with 1,000 images per class.
Matsuda et al.~\cite{Matsuda2012Multiple} constructed a Japanese food dataset UEC Food100 with 15K images in 100 dish categories. In comparison, ISIA Food500~\cite{Min-ISIA-500-MM2020} contains nearly 400k food images in 500 categories, which is the largest food image recognition. In addition, there are also recipe-related datasets. Salvador et al.~\cite{salvador2017learning} built the Recipe1M, with nearly 900k images and 1 million recipes, which is widely used in multi-modal learning between images and recipes. Based on Recipe1M, an even larger dataset Recipe1M+~\cite{marin2019learning} was constructed with more than 13 millions of food images. However, these datasets are mainly built to support food recognition and recipe generation research rather than food image segmentation, so they do not segment food images into multiple masks and labels of ingredient . UECFoodPix~\cite{uecfoodpix} and UECFoodPixComplete~\cite{uecfoodpixcomplete} are the only two datasets for food image segmentation, which contains 10,000 images with more than 100 categories. Nevertheless, their annotation are limited to dish-wise masks  so they cannot be used for ingredient segmentation. 

In this paper, we built FoodSeg103 dataset with 7,118 images and more than 40k masks covering 103 food ingredients. In addition, we have collected another image set for Asian food with 2,372 images (for cross-domain evaluation of the models). Combining the main set and the Asian set, we get the FoodSeg154 with nearly 10k images and 60k ingredient masks. To our best knowledge, FoodSeg154 is the first and the largest ingredient-level dataset for fine-grained food image segmentation. Dataset is a key step in developing deep learning based methods. We hope our dataset can inspire more efforts for the task of food image segmentation.

\begin{figure*}[t]
	\centering
	\subfigure[Easy cases]{
		\includegraphics[height=0.3\textwidth]{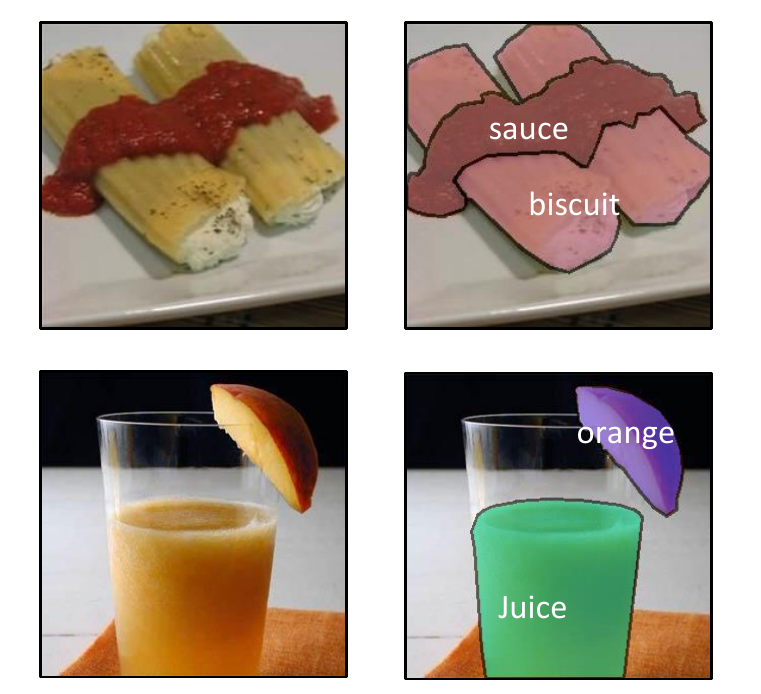}}
	\subfigure[Hard cases] {
		\includegraphics[height=0.3\textwidth]{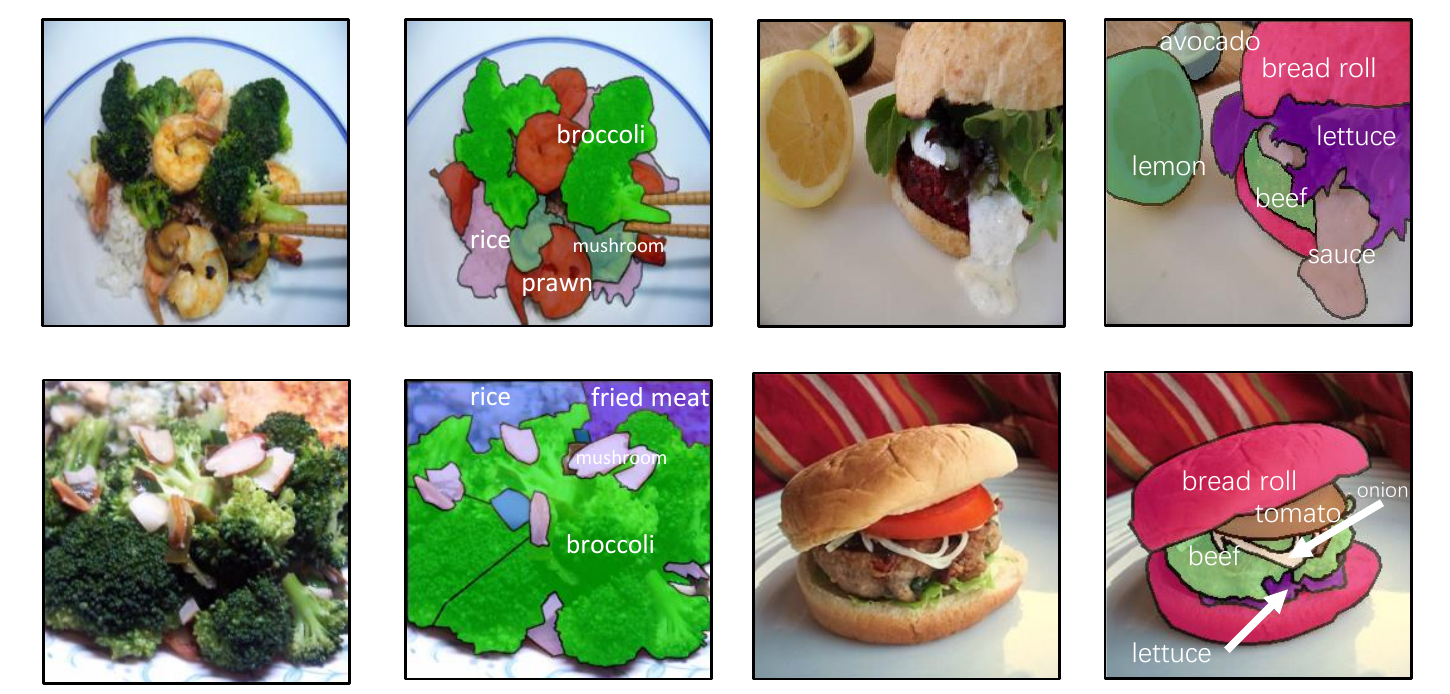}}
		\vspace{-3mm}
	\caption{Foodseg103 examples: source images (left) and annotations (right).} 	
	\label{fig:dataset}
\end{figure*}

\noindent{\textbf{Semantic Segmentation in Images.}} Deep learning based semantic segmentation is a super hot topic in recent years. 
Fully convolutional neural network (FCN)~\cite{long2015fully} is the first semantic segmentation framework based on deep convolutional neural networks. It predicts pixel-wise masks by replacing the fully connected layers with convolution layers and achieves a clear margin of improvement on the model performance. 
Chen et al.~\cite{chen2014semantic} proposed DeepLab which applies dilated convolutional layers in vanilla FCN. The trained model is more effective as the dilation mechanism enlarges the receptive fields while maintaining a high resolution in feature maps.
Chen et al~\cite{chen2017deeplab} proposed the DeepLab v2, which adds an ASPP module to integrate features of different dilation rates. 
To further include contextual cues, PSPNet~\cite{zhao2017pyramid} proposed a PPM module that aggregates the contextual information using different-size pooling layers. 
Wang et al.~\cite{wang2018non} proposed the non-local networks to encode the relationship between each pair of pixels in the feature map.
Based on the non-local networks, CCNet~\cite{huang2018ccnet} adopted a criss-cross attention layer to significantly economize the computation costs of calculating attentions.
Most recently, vision transformer (attention-based)~\cite{dosovitskiy2021an,NIPS2017_3f5ee243} was adapted to tackle semantic segmentation problems in~\cite{SETR}.
recently and achieves state-of-the-art results~\cite{SETR}.
In this paper, we conduct extensive experiments on our dataset using three representative semantic segmentation methods: CCNet~\cite{huang2018ccnet}, FPN~\cite{kirillov2019panoptic} and SeTR~\cite{SETR}. We also plug the proposed ReLeM into these methods to show its general efficiency.

\section{Food Image Segmentation Dataset}~\label{sec:fn}
FoodSeg103 is a subset of FoodSeg154, and the latter includes an additional subset of Asian food images and annotations.
Some example images and their annotations can be found in Figure~\ref{fig:dataset}.
In FoodSeg103, we have defined 103 ingredient categories and assigned these category labels as well as the segmentation masks to 7,118 images. The images are from an existing recipe dataset called Recipe1M~\cite{salvador2017learning}.
For the additional subset in FoodSeg154, we specially collect 2,372 images of Asian food which is of larger diversity than the Western food in FoodSeg103.
We use this subset to evaluate the domain adaptation performance of our food image segmentation models. \textbf{We release FoodSeg103 to facilitate public research, but currently we cannot make the Asian food set public due to the confidentiality of the images.}

\begin{figure*}[t]
	\centering
	\subfigure[Statistics of fine-grained ingredient categories (partial)]{
		\includegraphics[width=0.48\textwidth]{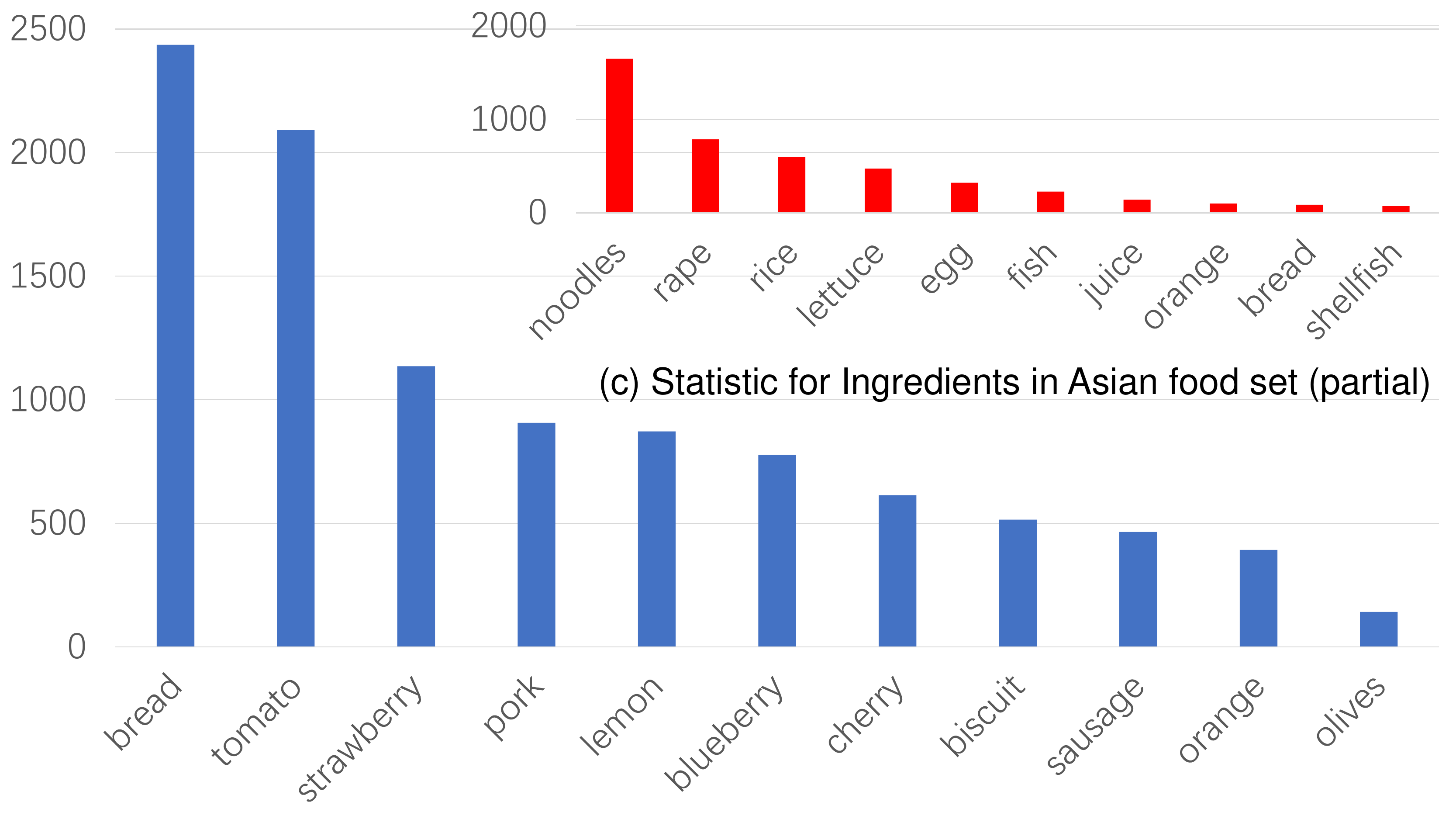}}
	\subfigure[Statistics of 15 super classes] {
		\includegraphics[width=0.48\textwidth]{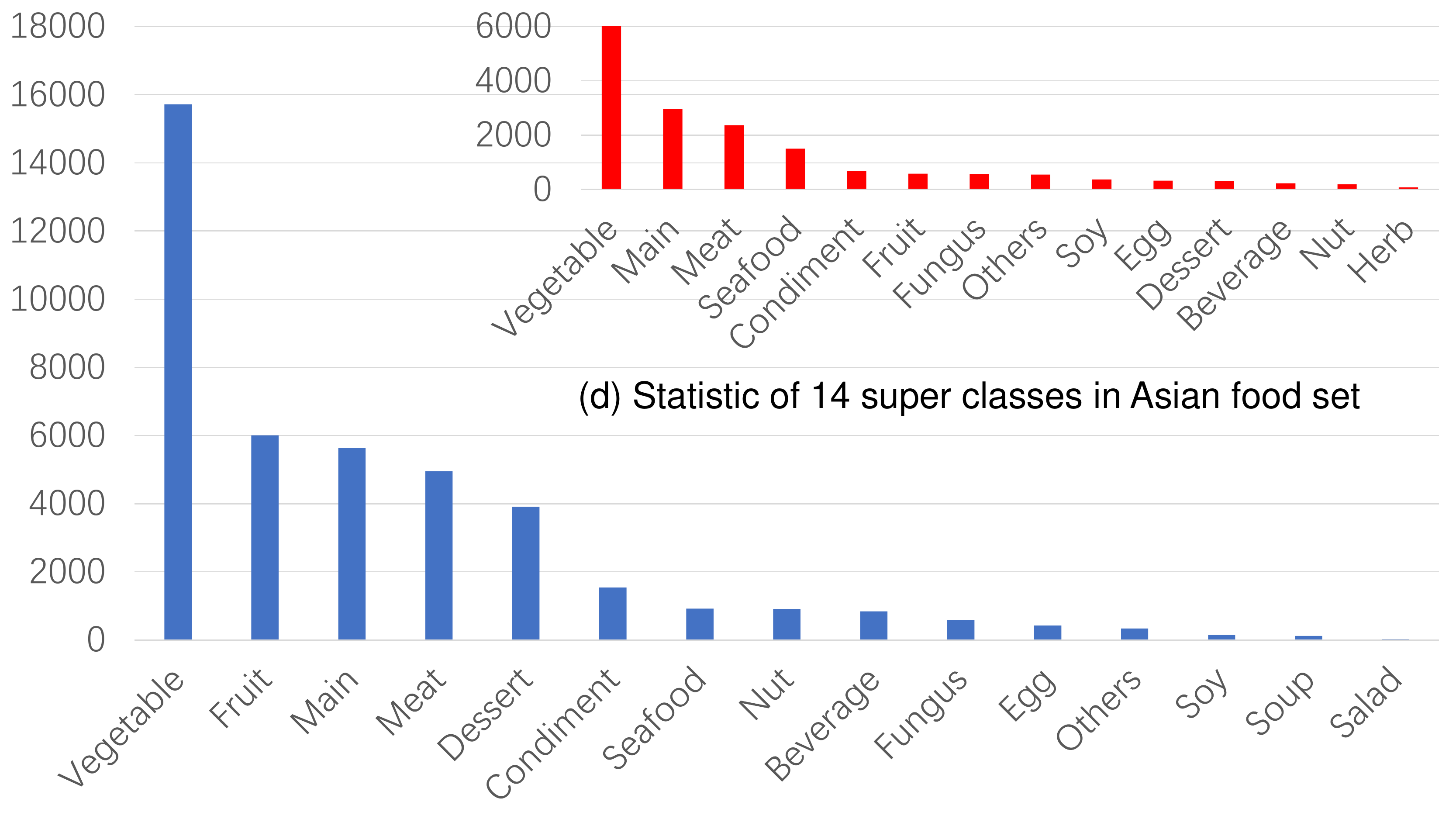}}
	\caption{
	Category statistics for our FoodSeg103 dataset in (a) and (b), and the Asian food image set (i.e., the additional set in FoodSeg154) in (c) and (d). 
	}\label{fig:distribution}
\end{figure*}

\subsection{Collecting Food Images}
We use FoodSeg103 as an example to elaborate the dataset construction process.
We elaborate the image source, category compilation and image selection as follows.
\textbf{Source:} We used Recipe1M~\cite{salvador2017learning, marin2019learning} as our source dataset.
This dataset contains 900k images with cooking instructions and ingredient labels, which are used for food image retrieval and recipe generation tasks.
\textbf{Categories:} First, we counted the frequency of all ingredient categories in Recipe1M. While there are around 1.5k ingredient categories~\cite{Salvador2019inversecooking}, most of them are not easy to be masked out from images.
Hence, we kept only the top 124 ingredient categories (with further refinement, this number became 103) and assigned ingredients with the ``others'' category when they do not fall under the above 124 categories.
Finally, we grouped these categories into 14 superclass categories, e.g., ``Main'' (i.e., main staple) is a superclass category covering more fine-grained categories such as ``noodle'' and ``rice''.
\textbf{Images:} In each fine-grained ingredient category, we sampled Recipe1M images based on the following two criteria: 
1) the image should contain at least two ingredients (with the same or different categories) but not more than 16 ingredients; and
2) the ingredients should be visible in the images and easy-to-annotate. 
Finally, we obtained 7,118 images to annotate masks.

\begin{figure}[t]
	\centering
	\subfigure[Source Images]{
		\includegraphics[width=0.15\textwidth]{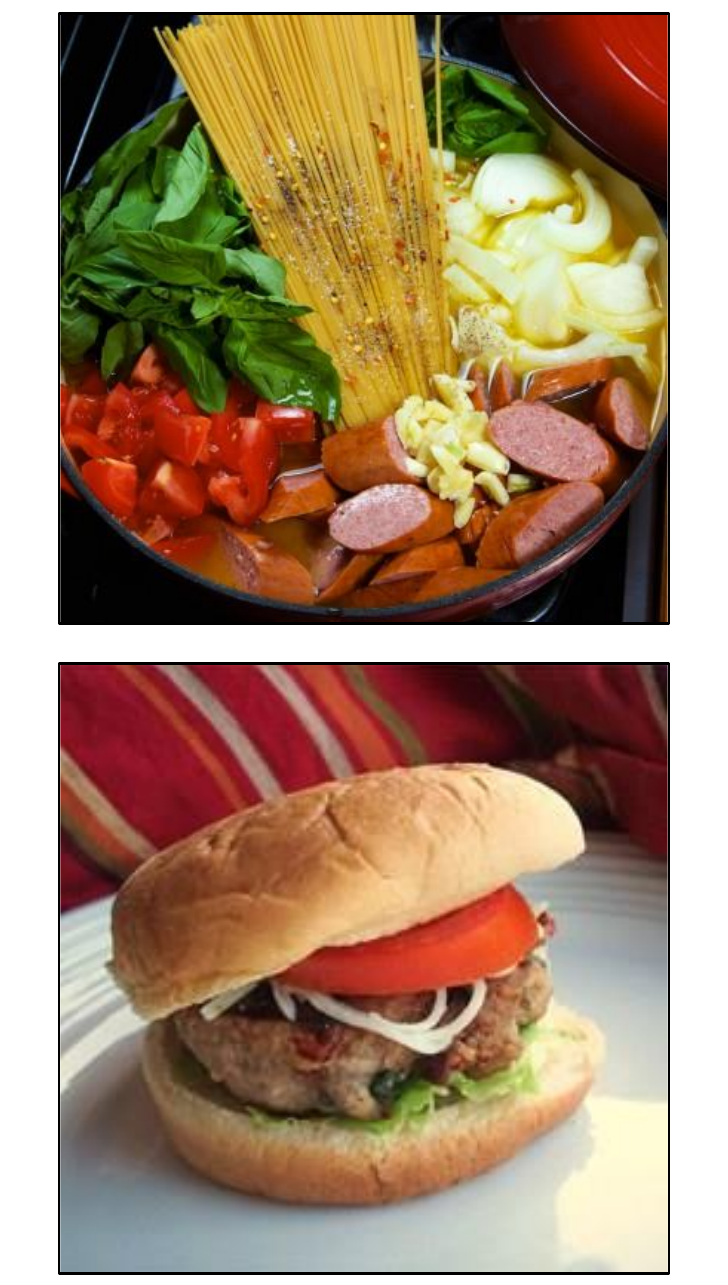}}
	\subfigure[Ingredient-level Annotation] {
		\includegraphics[width=0.15\textwidth]{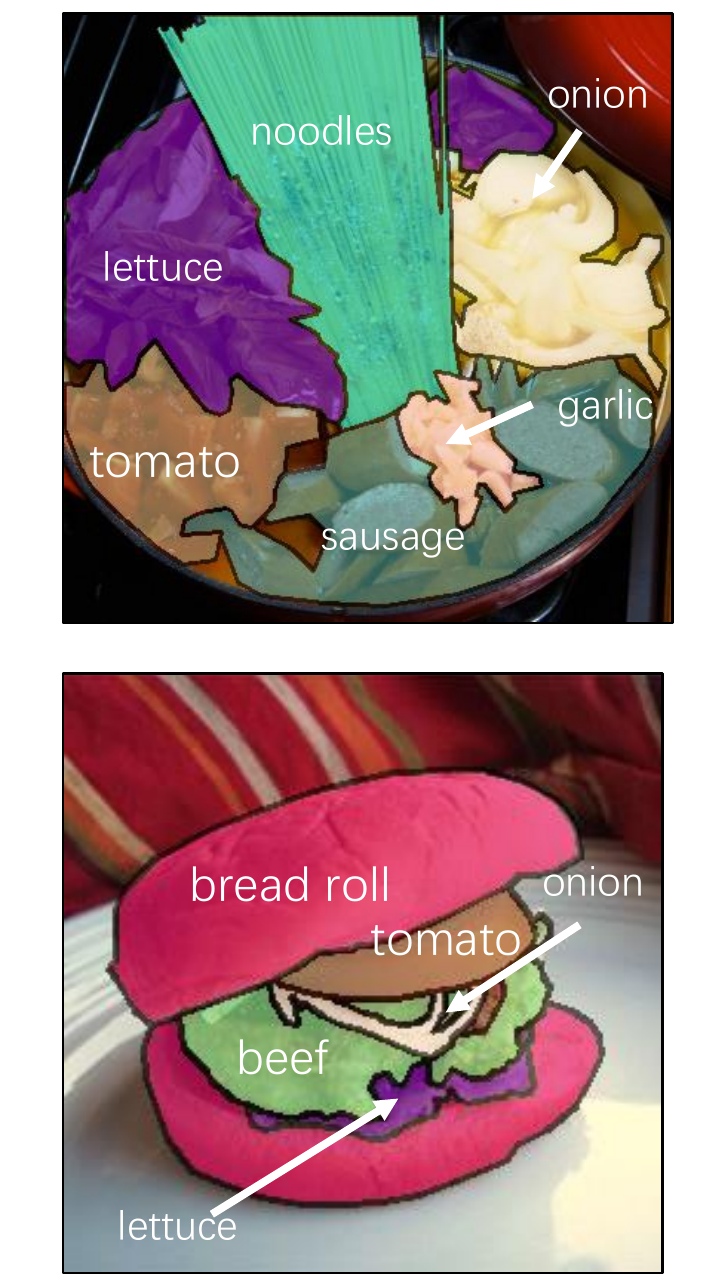}}
	\subfigure[Dish-level Annotation] {
		\includegraphics[width=0.15\textwidth]{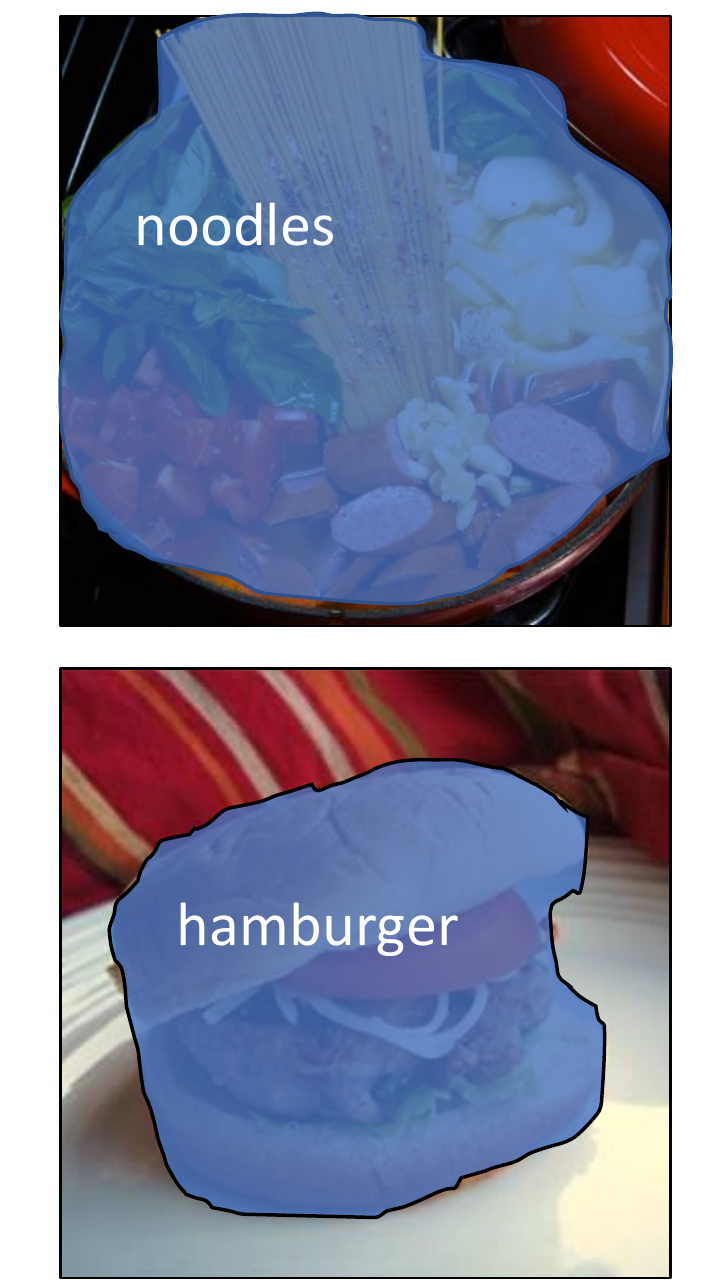}}\vspace{-2mm}
	\caption{Comparison of different annotation styles for masking food images: (a) source images, and (b) ingredient-level annotation (ours), and (c) dish-level annotation~\cite{uecfoodpixcomplete}. Ingredient-level annotation contains more details.} 	\label{fig:uec}
\end{figure}

\subsection{Annotating Ingredient Labels and Masks} \label{sec:collect}
\begin{figure}[htp]
	\centering
	\subfigure[Source Images]{
		\includegraphics[width=0.15\textwidth]{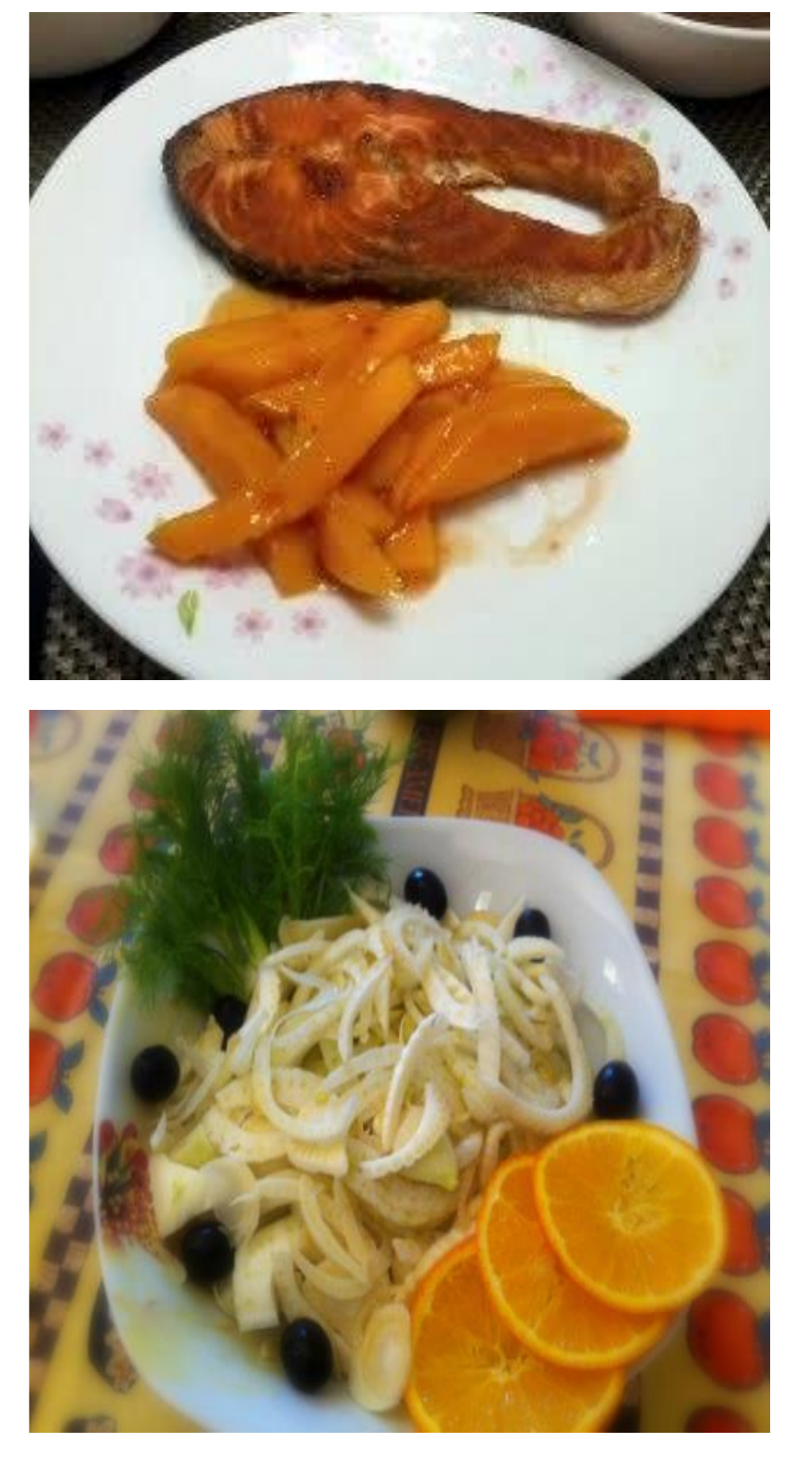}}
	\subfigure[Before Refinement] {
		\includegraphics[width=0.15\textwidth]{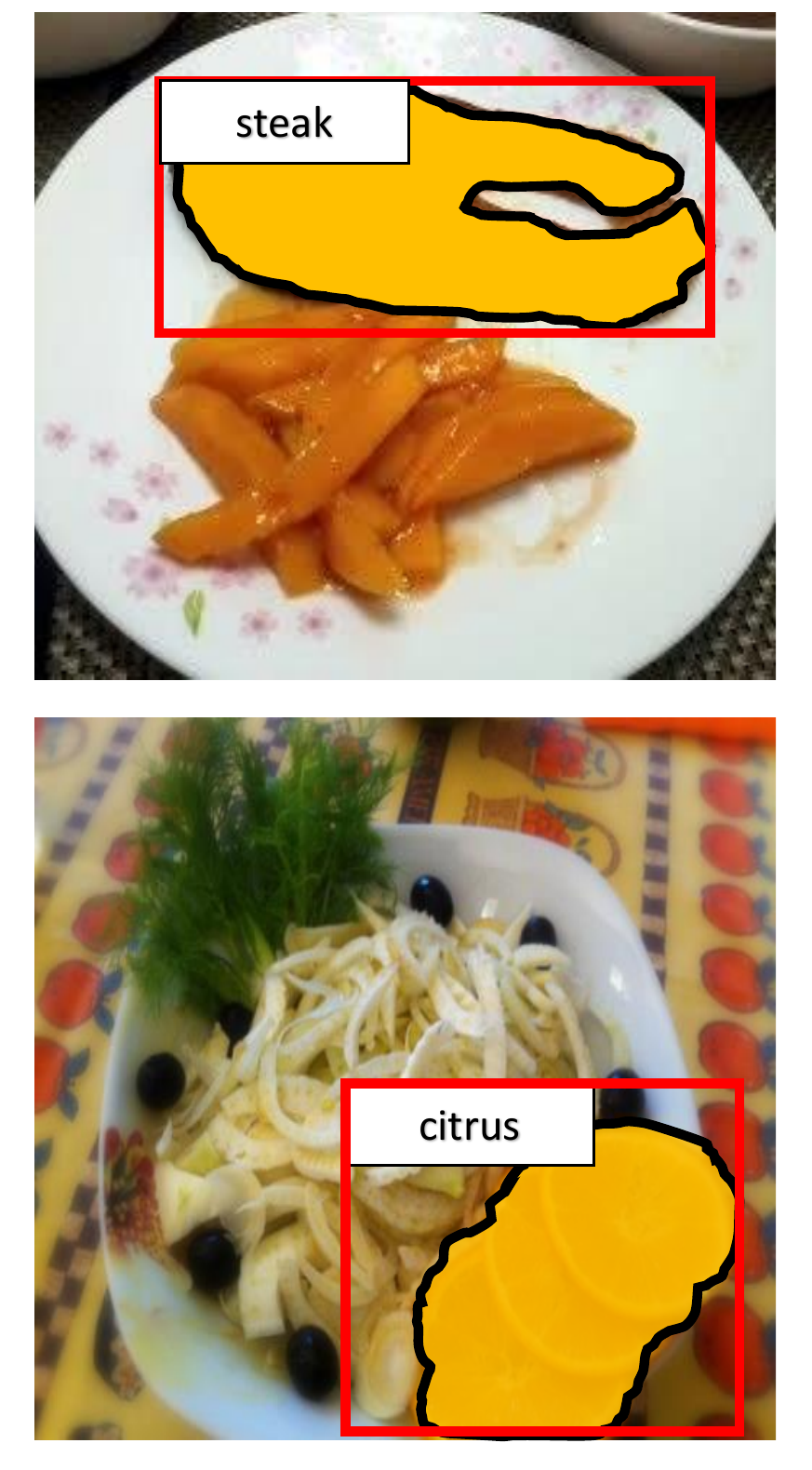}}
	\subfigure[After Refinement] {
		\includegraphics[width=0.15\textwidth]{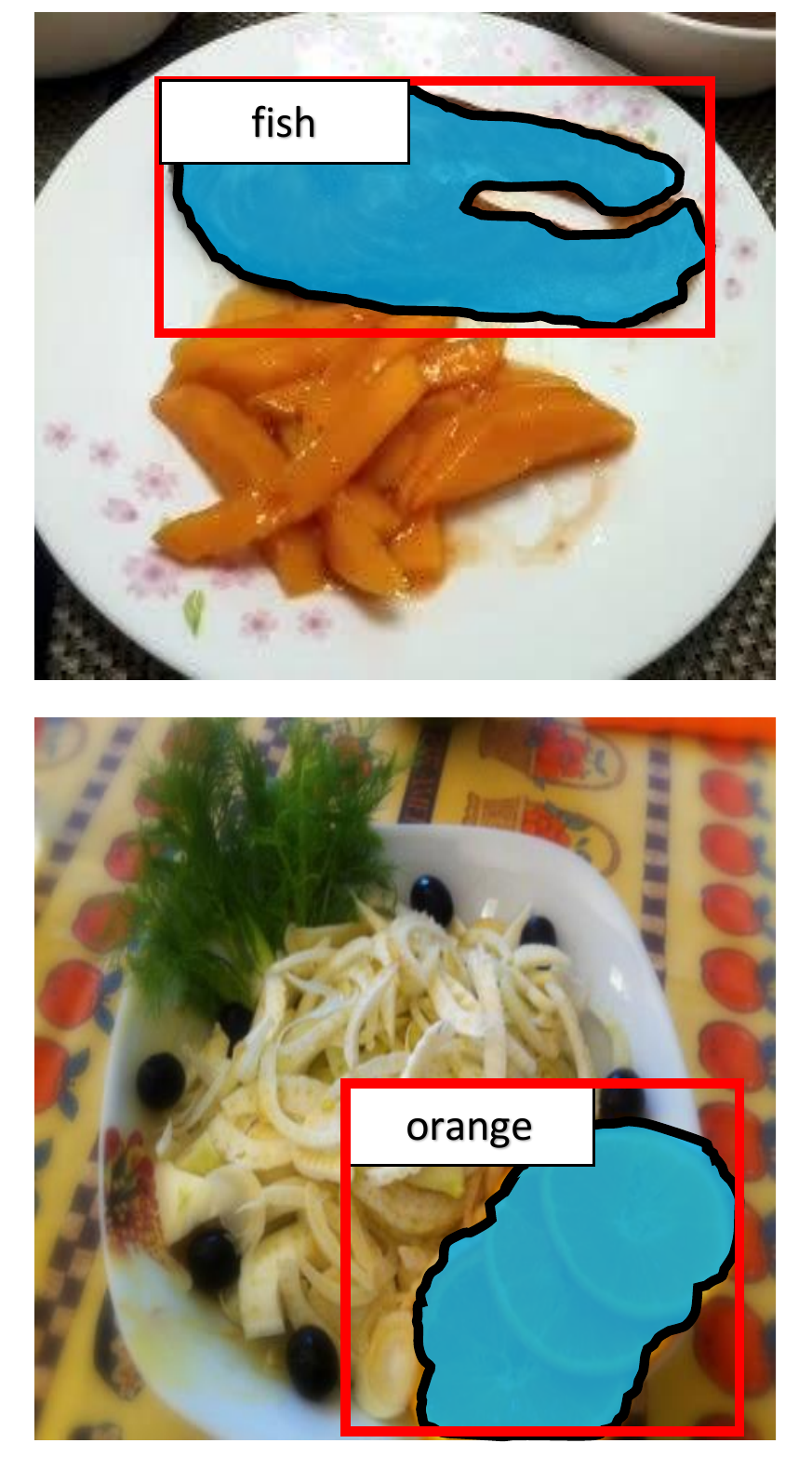}}
	\caption{Examples of dataset refinement. (a) sources images (b) before refinement (wrong or confusing labels exist), and (c) after refinement.} 	\label{fig:correct}
\end{figure}

Given the above images, the next step is to annotate segmentation masks, i.e., the polygons covering the pixel-wise locations of different ingredients.  This effort includes the mask annotation and mask refinement steps. 
\textbf{Annotation:} We engaged a data annotation company to perform mask annotation, a laborious and painstaking job.
For each image, a human annotator first identifies the categories of ingredients in the image, tags each ingredient with the appropriate category label and draws the pixel-wise mask. 
We asked the annotators to ignore tiny image regions (even if it may contain some ingredients) with area covering less than 5\% of the whole image.
\textbf{Refinement:} 
After receiving all masks from the annotation company, we further conducted an overall refinement.
We followed three refinement criteria: 1) correcting mislabeled data; 2) deleting unpopular category labels that are assigned to less than 5 images, and 3) merging visually similar ingredient categories, such as orange and citrus. After refinement, we reduced the initial set of 125 ingredient categories to 103.
Figure~\ref{fig:correct} shows some examples refined by us. 
The annotation and refinement works took around one year.

We show some data examples in Figure~\ref{fig:dataset}.
In Figure~\ref{fig:dataset} (a), we give some easy cases where the boundaries of ingredients are clear and the image compositions are not complex. 
In Figure~\ref{fig:dataset}~(b) and (c), we show some difficult cases with overlapped ingredient regions and complex compositions in the images.
Figure~\ref{fig:distribution} shows the distributions of fine-grained ingredient categories and superclass categories.
Figures~\ref{fig:distribution}(a) and \ref{fig:distribution}(c) show partial statistics for small subsets of categories due to page limit. The complete statistics will be published when releasing the dataset.

\subsection{Comparing with Food Image Datasets} \label{sec:comp}
\noindent{\textbf{Food Image Datasets}}. 
We summarize the comparison results in Table \ref{tab:compare}. 
We only include datasets that are mainly used for 
food recognition tasks.
They contain images and dish-level labels, and therefore they do not have any ingredient-level annotations. 
Recipe1M and Recipe1M+ include ingredient labels for each images but not the segmentation masks.
Notably, there are two datasets for food image segmentation: UECFoodPix~\cite{uecfoodpix} and UECFoodPixComplete~\cite{uecfoodpixcomplete}. Below, we compare these two with our datasets FoodSeg103 and FoodSeg154 in detail.

\noindent{\textbf{Food Image Segmentation Datasets}}. UECFoodPix and UECFoodPixComplete (UECFoodPixComp.) are two public datasets for food image segmentation, with 10k images and 102 dish categories. Detailed comparison numbers are given in Table \ref{tab:stat}. We highlight three advantages of our FoodSeg103 and FoodSeg154: 1) the number of pixel-wise masks of FoodSeg (40k and 60k) is significantly larger than UEC dataset (only 10k); 2) the annotation mask in UECFoodPix and UECFoodPixComp covers entire dish but not ingredients (dish components), while our FoodSeg154 and FoodSeg103 have ingredient-wise masks, which better capture the characteristic of the food.
Illustrative comparisons are given in Figure \ref{fig:uec}. 

In Table \ref{tab:stat}, we not only present the statistic numbers but also evaluate FoodSeg103, UECFoodPix and UECFoodPixComplete using deeplabv3+ as a baseline model.
The last row of the table shows that FoodSeg103 serves as a more challenging benchmark for semantic segmentation. Moreover, fine-grained ingredient annotations in our datasets are more useful for analyzing food nutrition and estimating calories in health-related applications.

\begin{table}[thp]
		\centering
		\resizebox{0.48\textwidth}{!}{
		\begin{tabular}{|l|c|c|c|c|c|}
			\hline \hline
			Dataset\;\; & Year & Type & \#Dish& \#Ingr. &Images \\
			\hline
			PFID~\cite{Chen2009PFID} & 2009 & CLS & 101&0& 4,545\\
			Food50~\cite{Joutou2010A} &2010&CLS & 50&0& 5,000\\
			Food85~\cite{Hoashi2010Image} &2010&CLS & 85& 0&5,500\\
			UEC Food100~\cite{Matsuda2012Multiple} &2012&CLS & 100&0& 14,361\\
			UEC Food256~\cite{kawano2014automatic}&2014&CLS  &256 &0&25,088\\
			ETH Food-101~\cite{Bossard-Food101-ECCV2014}&2014&CLS  &101 &0&101,000\\
			UPMC Food-101~\cite{Wang-RRLMFD-ICME2015}&2015&CLS &101&0& 90,840\\
			Geo-Dish~\cite{xu2015geolocalized}&2015&CLS &701&0&117,504\\
			Sushi-50~\cite{Qiu-MDFR-BMVC2019}&2019&CLS&50&0&3,963\\
			FoodX-251~\cite{Parneet-FoodX251-CVPRW2019}&2019&CLS&251&0&158,846\\
			ISIA Food-200~\cite{Min-IGCMAN-ACMMM2019}&2019&CLS&200&0&197,323\\
			FoodAI-756~\cite{Doyen-FoodAI-KDD2019}&2019&CLS&756&0&400,000\\
			Recipe1M~\cite{salvador2017learning} & 2017 & Recipe & 0 &1488& 1M\\
			Recipe1M+~\cite{marin2019learning} & 2019 & Recipe & 0 &1488& 14M\\
			UECFoodPix~\cite{uecfoodpix} & 2019 & SEG & 102 &0& 10,000\\
			UECFoodPixComp.~\cite{uecfoodpixcomplete} & 2020 & SEG & 102& 0& 10,000\\
			\hline
			FoodSeg103&2021&SEG&730 &103&7,118 \\
			FoodSeg154&2021&SEG&730 &154&9,490\\
			\hline
			\hline
		\end{tabular}}
		\caption{A global view of existing food image datasets. (CLS: no recipe and masks, Recipe: with recipe, SEG: with segmentation masks ) 
		}\label{tab:compare}
	\end{table}\vspace{-1mm}

\begin{table}[thp]
		\centering
		\resizebox{0.48\textwidth}{!}
		{
		\begin{tabular}{|l|c|c|c|c|}
			\hline 
			Statistics\;\; & FoodSeg103 & FoodSeg154&UECFood &UECFoodComp. \\
			\hline
			\# Dish & 730 & 730&102&102\\
			\# Ingr. & 103 & 154&0&0\\
			\# images & 7,118& 9,490&10,000& 10,000\\
			\# masks & 42,097& 59,773&14,011&16,060\\
			mean image width & 771 pixels&776 pixels &442 pixels&442 pixels\\
			mean image height & 647 pixels&656 pixels &  349 pixels&  349 pixels\\
			mIoU@deeplabv3+& 34.2 & N.A. &41.6 &  55.5 \\
			\hline
			\hline
		\end{tabular}}\vspace{2.5mm}
		\caption{Data summary and comparison with existing food image segmentation datasets.
		}\label{tab:stat}
	\end{table}

\section{Food Image Segmentation Framework}~\label{sec:dl}
\begin{figure}[ht]
\begin{center}
\includegraphics[width=1.0\linewidth]{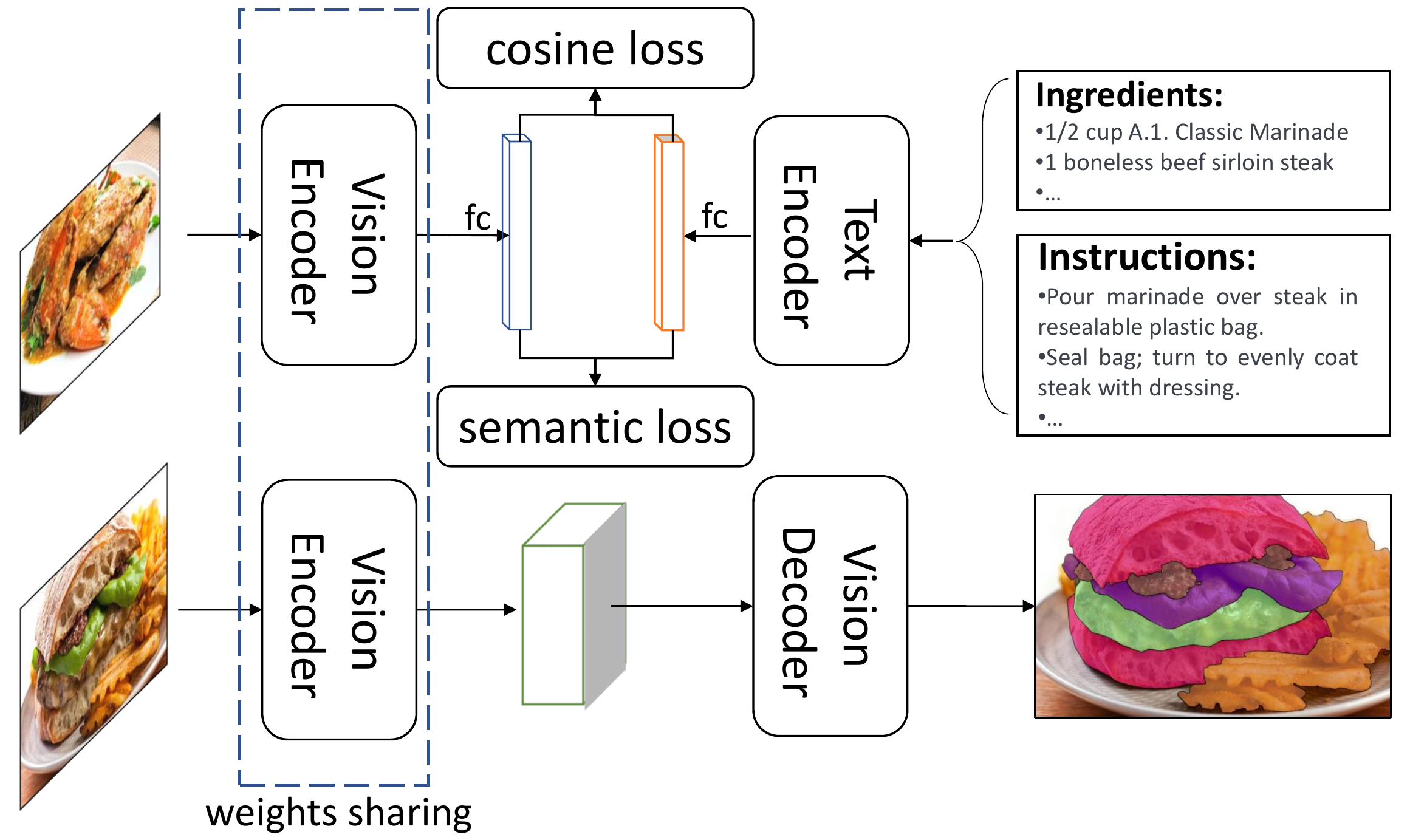}
\end{center}
\caption{Our food image segmentation framework consists of two modules: Recipe Learning Module (ReLeM) and Image Segmentation Module (Segmenter). 
For ReLeM, we encode the recipe information into the visual representation of the food image. We deploy the cosine similarity to compute the distance between two distinct-modality models, together with a semantic loss~\cite{salvador2017learning}.
After training, we use the trained encoder to initialize the encoder of the Segmenter.
The decoder of the Segmenter is trained with the segmentation masks from a random initialization.}
\label{fig:pipeline}
\end{figure}

As shown in Figure~\ref{fig:pipeline},
our food image segmentation framework contains two modules.
One is the \textit{recipe learning module} (ReLeM) to incorporate recipes in the form of language embedding into the visual representation of a food image.
We call this approach multi-modality knowledge transfer.
In this approach, we explicitly force the visual representations of the same ingredient appearing in different dishes to be ``connected'' in the feature space through the common language embedding (extracted from the ingredient label and its cooking instructions), so as to handle the high variance of the ingredient appearing in different dishes.
The other module of our framework is the \textit{encoder-decoder based image segmentation}. Its encoder is initialized using the one trained by ReLeM, and its decoder is randomly initialized and trained with the segmentation masks.
We next introduce the two modules in detail.

\begin{figure}[htp]
	\centering
	\subfigure[Source Image]{
		\includegraphics[height=0.13\textwidth]{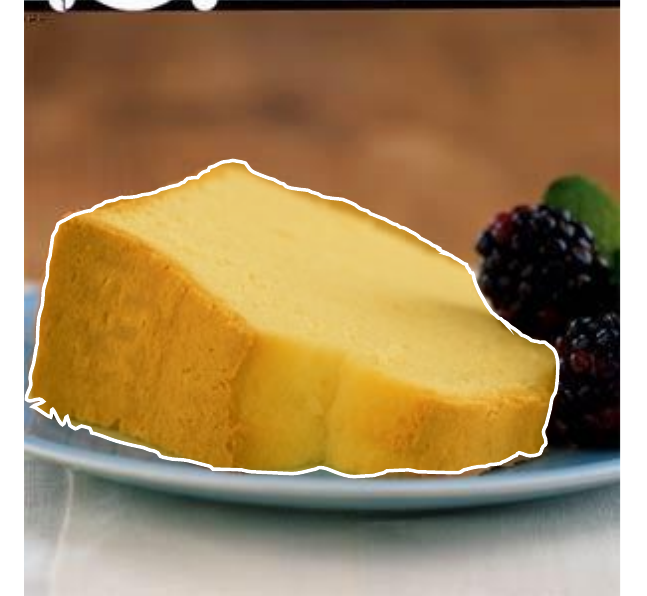}}
	\subfigure[Mask Prediction] {
		\includegraphics[height=0.13\textwidth]{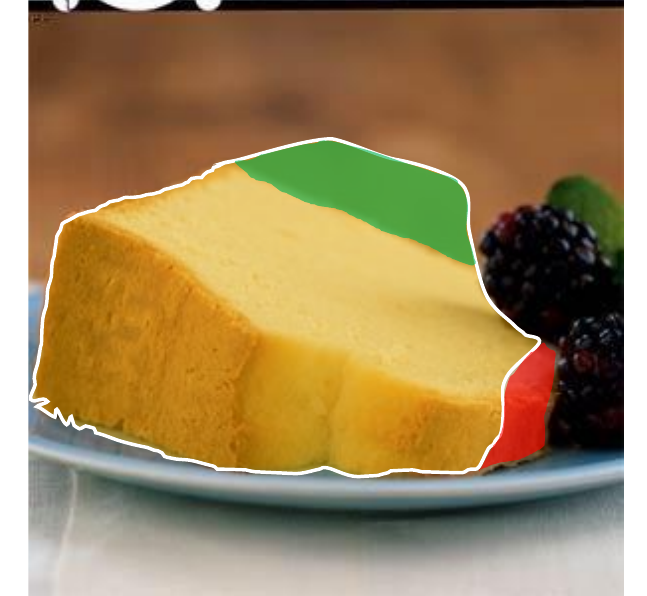}}
	\subfigure[Label Marks] {
		\includegraphics[height=0.13\textwidth]{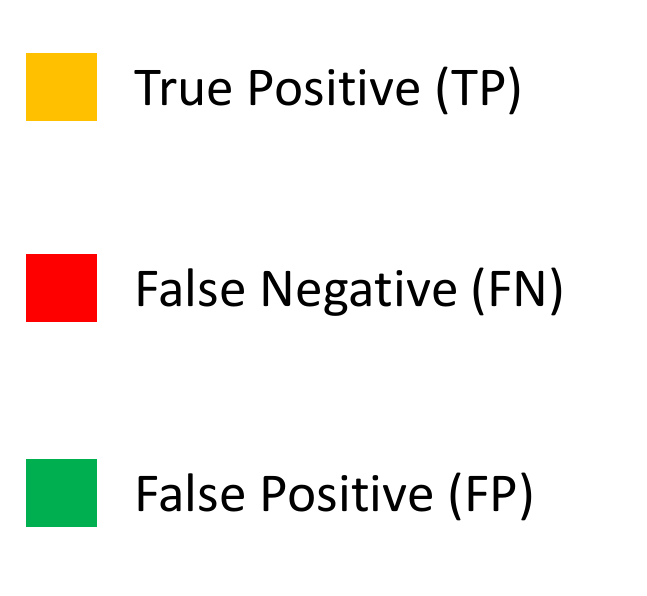}}
	\caption{Calculating IoU and Acc, taking the ``cake'' mask as an example. IoU = ($\frac{\text{TP}}{\text{TP+FP+FN}}$) and Acc = ($\frac{\text{TP}}{\text{TP+FN}}$).} 	\label{fig:iou}
\end{figure}

Food image segmentation can be viewed as a special type of semantic segmentation~\cite{lin2016fpn, SETR}.
It is more difficult than normal image segmentation due to: 1) the ingredient cooked with different methods can vary a lot by appearances, and 2) ingredient distribution is inevitably long-tailed making the data very sparse for ingredients in the long tail.
Given a food image, the Segmenter identifies the ingredient categories and also mask out the corresponding pixels for each category (class).
The common metrics for measuring Segmenter's performance include mIoU (mean IoU over each class), mACC (mean accuracy over all classes) and aAcc (over all pixels), See Figure \ref{fig:iou} for more details of IoU and accuracy (Acc) calculation. 

\subsection{Recipe Learning Module (ReLeM)}
\noindent{\textbf{Overview.}}
We propose ReLeM to reduce the large intra-variance of ingredients caused by different cooking methods mentioned in the recipes.
Specifically, our training method integrates the recipe information into the visual representation of the corresponding image. 
Assume an ingredient in two different images are cooked in different methods.
The visual representations of the ingredients from vision encoder are denoted as $v_1$ and $v_2$, where $v_1$ and $v_2$ have significant difference in the visual space. 
ReLeM aims to reduce this difference according to its word embedding of the cooking instructions of the two recipes $r_1$ and $r_2$ respectively in the language space.
\begin{equation}
    |\phi(v_1|r_1) -\phi(v_2|r_2)| < |\phi(v_1)- \phi(v_2)|
\end{equation}
where $\phi$ is the vision decoder in the Segmenter (elaborated in Section~\ref{sec:vs}). 

Our ReLeM is optimized by using two loss terms: cosine similarity loss between features, and semantic loss (distance) between the text representation $t$ and the visual representation $v$ of the same image:
\begin{equation}
  L_{\text{cosine}}((v,t),y) = \left\{\begin{matrix}
  1-cosine(v,t)& y= 1\\ 
max(0,cosine(v,t)-\alpha) & y= -1
\end{matrix}\right.
\end{equation}
\begin{equation}
  L_{\text{semantic}}((v,t), u_v, u_t) = \text{CE}(v, u_v) + \text{CE}(t, u_t) 
\end{equation}
where $y$ denotes whether $t$ and $v$ are from the same recipe. $u_v$ and $u_t$ denote the semantic class of $u$ and $v$ respectively, and $\alpha$ is the margin parameter, which is set to 0.1.
As Recipe1M does not contain specific semantic labels (i.e., dish names), we define 2,000 semantic labels for it by selecting the most frequent dish names appeared in its recipe titles.

\noindent{\textbf{Preprocessing.}} Each recipe contains ingredients and cooking instructions. Some preprocessing steps are required to encode ingredients and instructions from raw text into the fixed length vectors before they are fed into the text encoder. 
Specifically,
we first extract useful ingredient and instruction texts from the raw recipe data by removing redundant words. 
For each ingredient, we learn a word2vec~\cite{mikolov2013efficient} representation using a bi-directional LSTM.
As the sequence of instructions can be long, it is difficult for LSTM to encode them, due to the gradient vanishing issue. Following a previous work~\cite{salvador2017learning}, we encode the instructions with a skip-instructions~\cite{kiros2015skip} to generate the feature vectors with a fixed length.

\noindent{\textbf{Text Encoder.}} The text encoder is a general module to extract text knowledge from ingredient labels and cooking instructions. We use two types of text encoders: \textit{LSTM-based encoder} and \textit{transformer-based encoder}. For LSTM-based, we use a bi-directional LSTM to encode ingredient features and a LSTM to encode instruction features. For transformer-based model, we use two light-weight transformers, each of which contains 2 transformer layers with 4-head self-attention modules.

\noindent{\textbf{Vision Encoder.}} The vision encoder used in ReLeM aims to extract the visual knowledge from the input image, and the weights will initialize the vision encoder in the segmenter. In this paper, two vision encoders are used: ResNet-50~\cite{he2016deep} based on convolutional neural network and ViT-16/B~\cite{dosovitskiy2021an} based on vision transformers.

\subsection{Image Segmentation Module (Segmenter)} \label{sec:vs}
Our framework follows the standard paradigm of semantic segmentation, where the input image is first encoded in a vision encoder, and then goes through a vision decoder for mask prediction. 
The existing segmentation models can be roughly divided into three groups, based on the different designs of encoder and decoder: \textit{Dilation based}, \textit{Feature Pyramid Networks (FPN) based} and \textit{Transformer based}.

\begin{figure*}[htp]
	\centering
	\subfigure[Dilation Based ]{
		\includegraphics[height=0.14\textwidth]{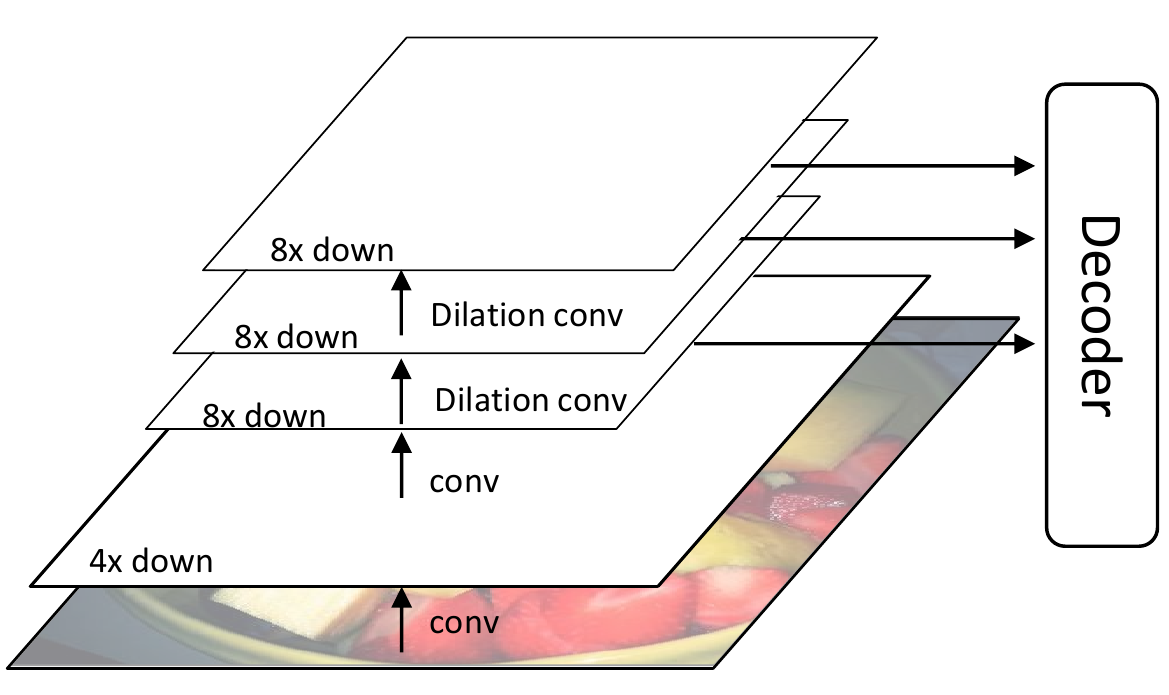}}
	\subfigure[FPN Based ] {
		\includegraphics[height=0.14\textwidth]{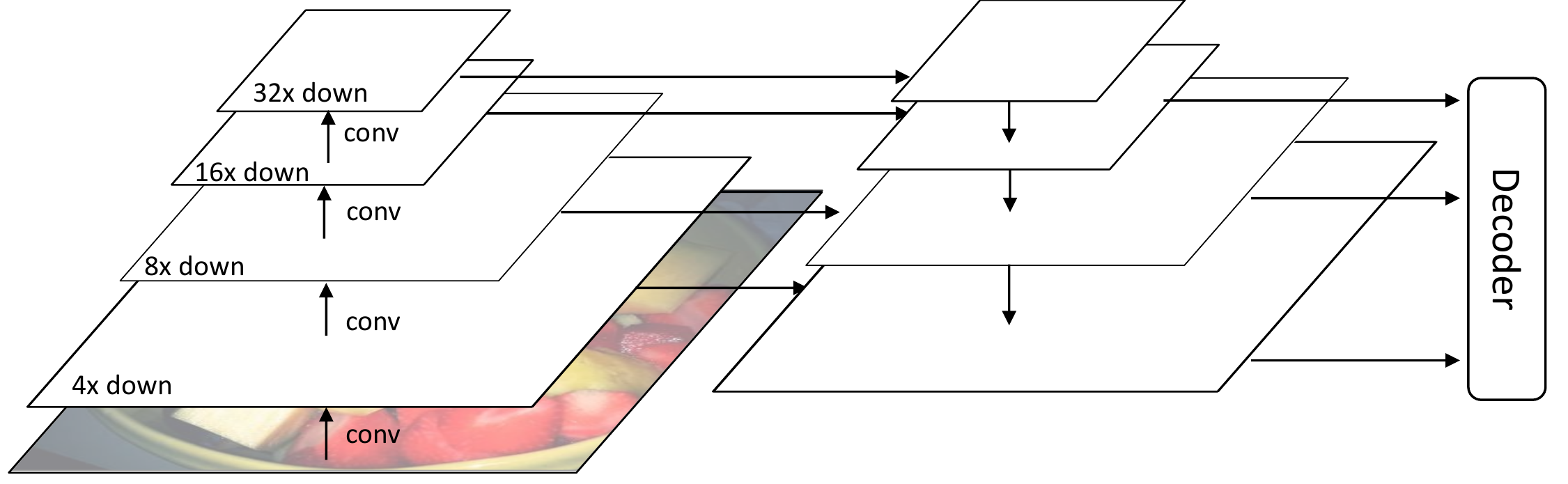}}
	\subfigure[Transformer Based ] {
		\includegraphics[height=0.14\textwidth]{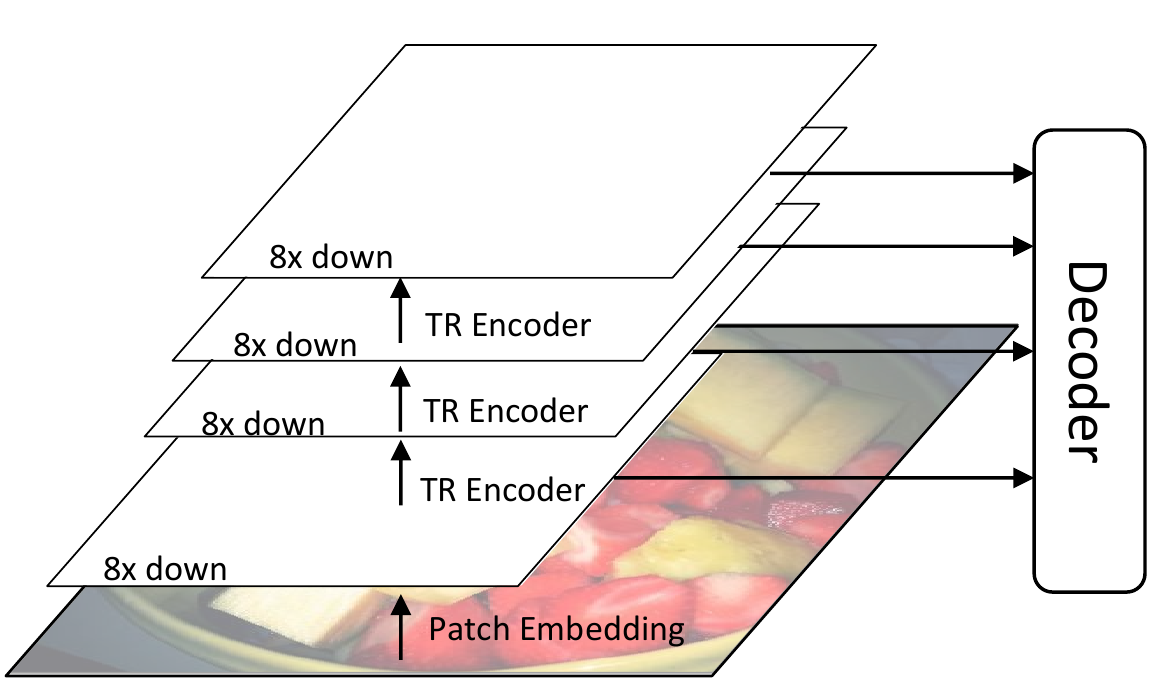}}
		\vspace{-1mm}
	\caption{Different types of encoder for food image segmentation} 	\label{fig:encoder}
\end{figure*}

\begin{figure*}[htp]
	\centering
	\subfigure[Dilation Based ]{
		\includegraphics[height=0.15\textwidth]{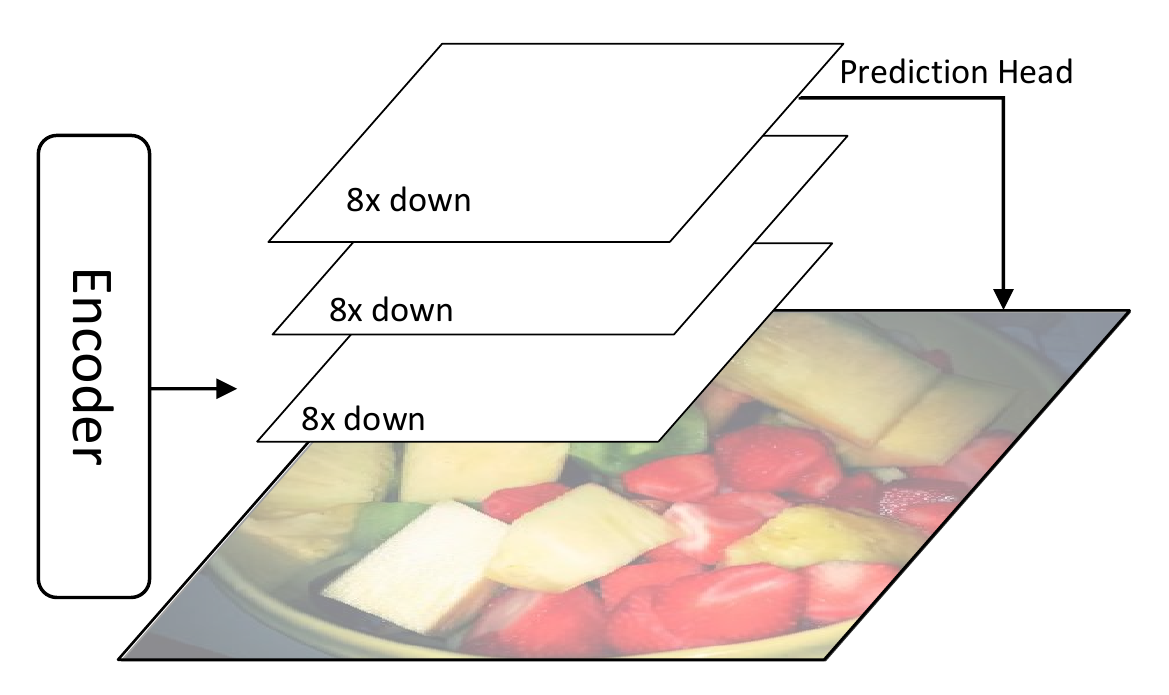}}
	\subfigure[FPN Based ] {
		\includegraphics[height=0.15\textwidth]{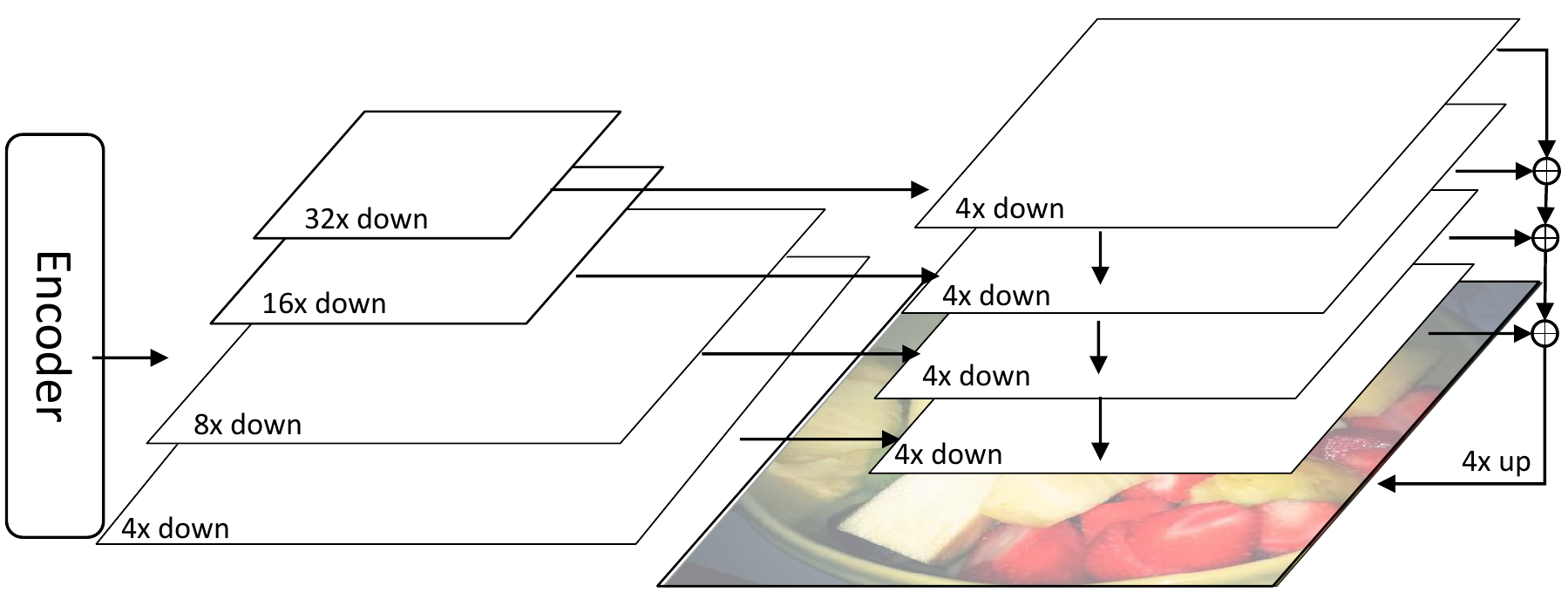}}
	\subfigure[Transformer Based ] {
		\includegraphics[height=0.15\textwidth]{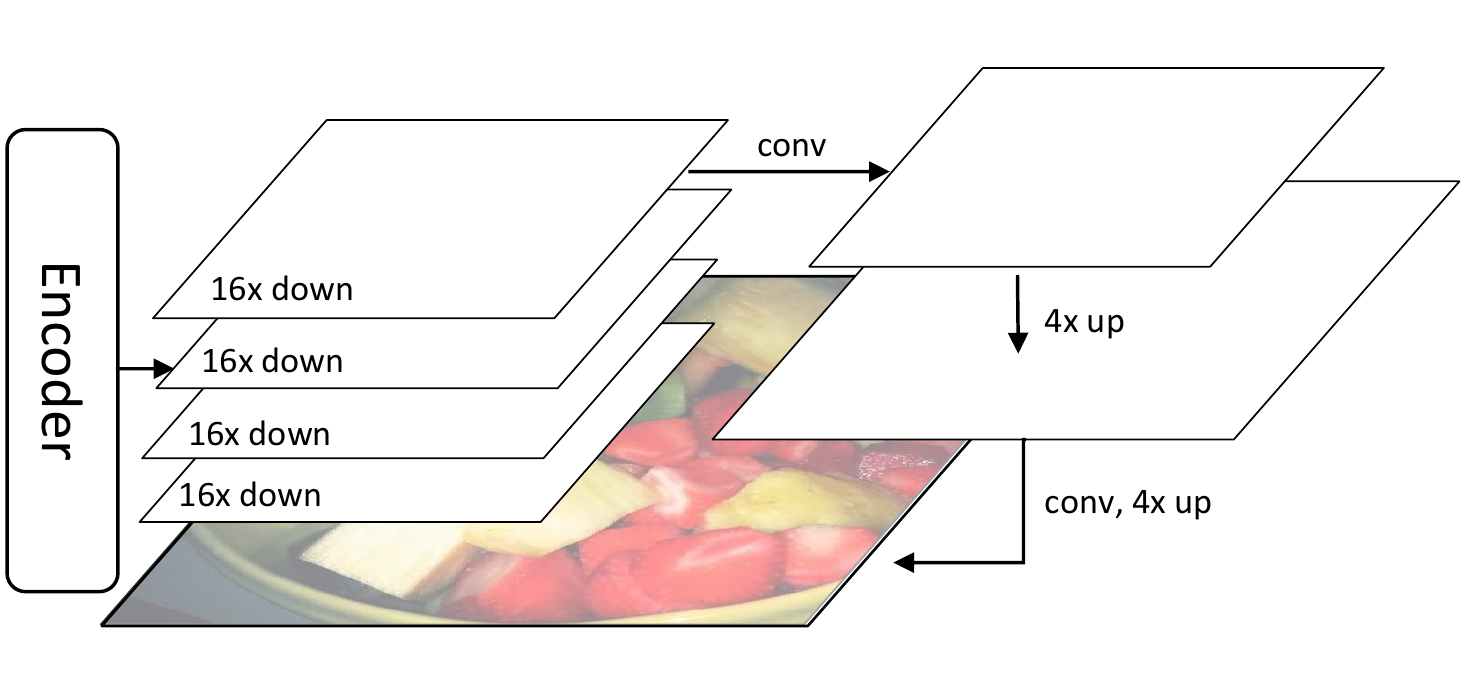}}\vspace{-1mm}
	\caption{Different types of decoder for food image segmentation} 	\label{fig:decoder}
\end{figure*}

\noindent\textbf{Dilation based}. 
Dilation convolution layers aim to enlarge the receptive fields without sacrificing the resolution, as shown in Figure~\ref{fig:encoder}~(a). 
In its decoder, only the last-layer feature maps are used for prediction~\cite{huang2018ccnet,chen2017deeplab}, as shown in Figure~\ref{fig:decoder}~(a). 

\noindent\textbf{FPN based}. 
FPN integrates feature maps in different layers by the lateral connection. The shallow-layer image representation is enhanced by integrating the feature maps generated in deep layers, as shown in  Figure~\ref{fig:encoder}~(b). 
In its decoder, a set of feature pyramids are merged together followed with a mask predictor, as shown in Figure~\ref{fig:decoder}~(b). 

\noindent\textbf{Transformer based}. Transformer is based on attention,
which suits semantic segmentation tasks well----the contextual information is important in segmenting objects.
Moreover, the receptive fields can be enlarged via attention mechanism~\cite{SETR,NIPS2017_3f5ee243}. 
The transformer-based model reshapes the image into a sequence of regions and then encodes them by a sequence of attention modules, as shown in Figure~\ref{fig:encoder}~(c). 
Its decoder predicts segmentation masks on the last-layer feature maps, as shown in Figure~\ref{fig:decoder}(c). 

In this paper, we conduct experiments using three representative frameworks of these three types, respectively, i.e., CCNet (Dilation)~\cite{huang2018ccnet}, FPN~\cite{kirillov2019panoptic} and SeTR (Transformer)~\cite{SETR}. 
Note that the encoder of Segmenter is pre-trained 
by our ReLeM.  With LSTM and transformer-based text encoding, we arrive at 6 different ReLeM models, i.e., 
ReLeM-$\{$ CCNet, FPN, SeTR$\} \times (\{$ LSTM, Transformer$\})$.
We use the standard pixel-wise cross-entropy loss to optimize segmentation models.

\section{Experiments}\label{sec:ep}

We conduct extensive experiments on our dataset FoodSeg103 and implement our proposed ReLeM by incorporating three baseline methods of semantic segmentation. 
Below, we first elaborate the experimental settings and the results of an ablation study.
Then, we show the performance gaps of the top model in the typical semantic segmentation task and our food image segmentation task.
We also evaluate the model adaptability using the Asian food data splits in our FoodSeg154.
Lastly, we provide some qualitative results of our best segmentation models.

\subsection{Implementation Details}
\noindent{\textbf{Dataset Settings}}  In our experiments, we use FoodSeg103 for in-domain training and testing, and use the additional Asian food set for out-domain testing. We randomly divide FoodSeg103 dataset into two splits: training set and testing set,  according to the 7:3 ratio. Our training set contains 4,983 images with 29,530 ingredient masks, while testing set contains 2,135 images with 12,567 ingredient masks. For ReLeM training, we use the training set of Recipe1M+ to learn the recipe representations (with test images in FoodSeg103 hidden from training). 

\noindent{\textbf{Segmenter Settings}} We conduct experiments based on two types of vision encoders: ResNet-50~\cite{he2016deep} based on convolutional neural networks, and ViT-16/B~\cite{dosovitskiy2021an} based on vision transformer. ResNet-50 is initialized from the pre-training model on ImageNet-1k~\cite{deng2009imagenet}, which is widely used in multiple vision tasks~\cite{krizhevsky2012imagenet,ren2015faster,chen2017deeplab}. ViT-16/B~\cite{dosovitskiy2021an} is a transformer-based model, which is initialized from the pre-training model on ImageNet-21k. ViT-16/B contains 12 transformer encoders with 12-head self-attention modules. We use the bilinear interpolation method to reinitialize the pre-trained positional embedding. In this paper, we use three types of segmentors: CCNet~\cite{huang2018ccnet}, FPN~\cite{kirillov2019panoptic} and SeTR~\cite{SETR}. CCNet and FPN are based on ResNet-50, while SeTR is based on ViT-16/B. 
Notably, SeTR extracts feature maps from $12^{\text{th}}$ transformer encoders, followed by two sets of convolution layers for prediction. Other components of the segmentors follow the default settings with random initialization.

\noindent{\textbf{ReLeM Settings}} We use two types of vision encoders in ReLeM: ResNet-50 and ViT-16/B, which follow the same setting as Segmenter. In text preprocessing step, we use the skip-instruction models from the pre-trained weights in \cite{Salvador-LCME-arXiv2018}.

\noindent{\textbf{Learning Parameters of Segmenter}} Each image will be resized into a fixed size of $2049 \times 1024$ pixels with a ratio range from 0.5 to 2.0. A $768 \times 768$ patch is cropped from the resized images, and random horizontal flipping and color jitter are applied. We trained the models with 80k iterations based on 8 images per batch, and optimized the models by SGD solvers, with a momentum as 0.9 and weight decay as 0.0005. For CCNet and FPN, we set the initial learning rate to 1e-3, while for SeTR we set initial learning rate to 1e-3. According to the general settings~\cite{wang2021pyramid,huang2018ccnet}, the learning rate is decayed by a power of 0.9 according to the polynomial decay schedule. For simplicity, we do not apply hard negative mining during training, and our framework is based on the widely used platform mmsegmentation~\cite{mmseg2020}. All experiments were conducted on 4 Tesla-V100 GPU cards.

\noindent{\textbf{Learning Parameters of ReLeM}} Each input image are resized into a size of $256 \times 256$ pixels and a $224 \times 224$ patch is cropped from the resized images as the input of the vision encoder. The model is trained for 720 epochs and each batch contains 160 images. We use Adam solver~\cite{kingma2014adam} to optimize the models, with a learning rate of 1e-4, Here we follow a two-stage optimization strategy. We first freeze the weights of the vision encoder and optimize the text encoder. After the text encoder converges, we start to train the vision encoder and freeze the parameters of the text encoder.

\begin{figure*}[t]
	\centering
	\subfigure[Source Image]{
		\includegraphics[width=0.18\textwidth]{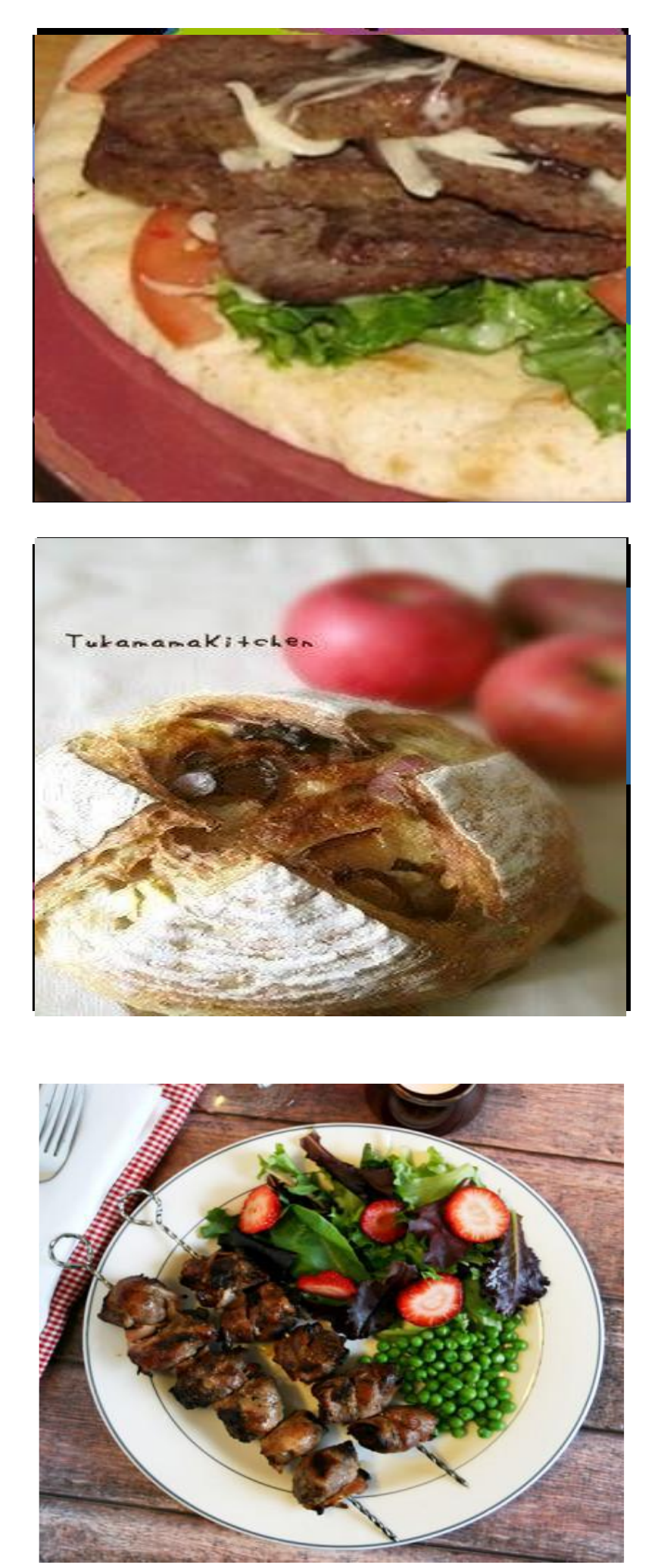}}
	\subfigure[Ground Truth] {
		\includegraphics[width=0.18\textwidth]{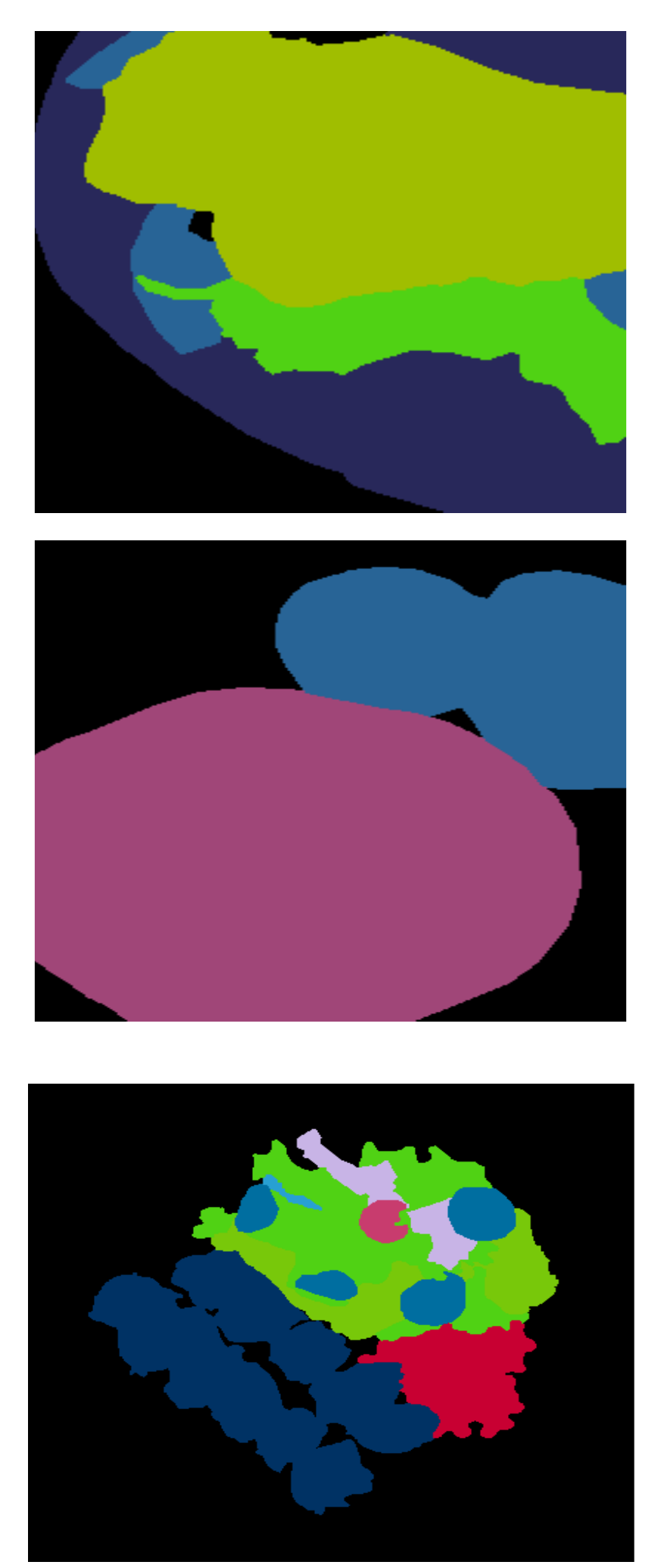}}
	\subfigure[ReLeM-CCNet] {
		\includegraphics[width=0.18\textwidth]{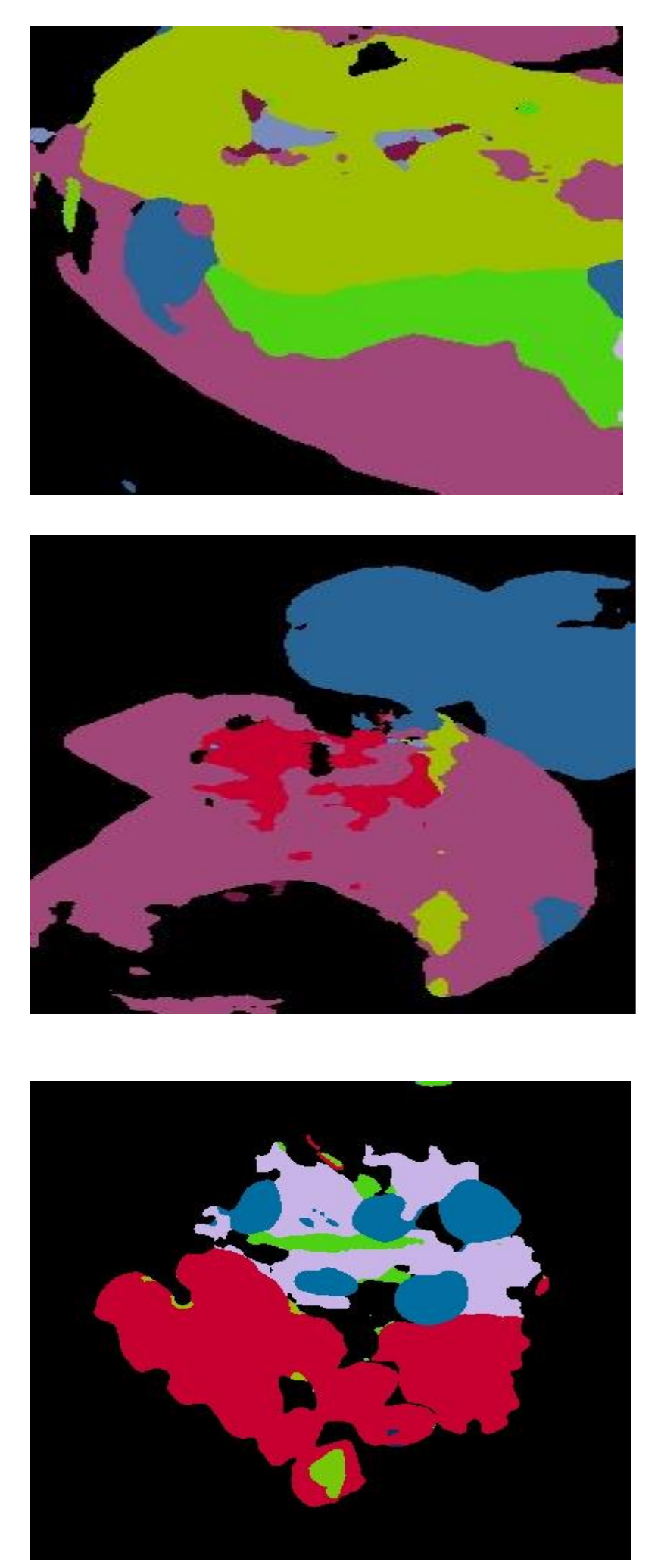}}
	\subfigure[CCNet] {
		\includegraphics[width=0.18\textwidth]{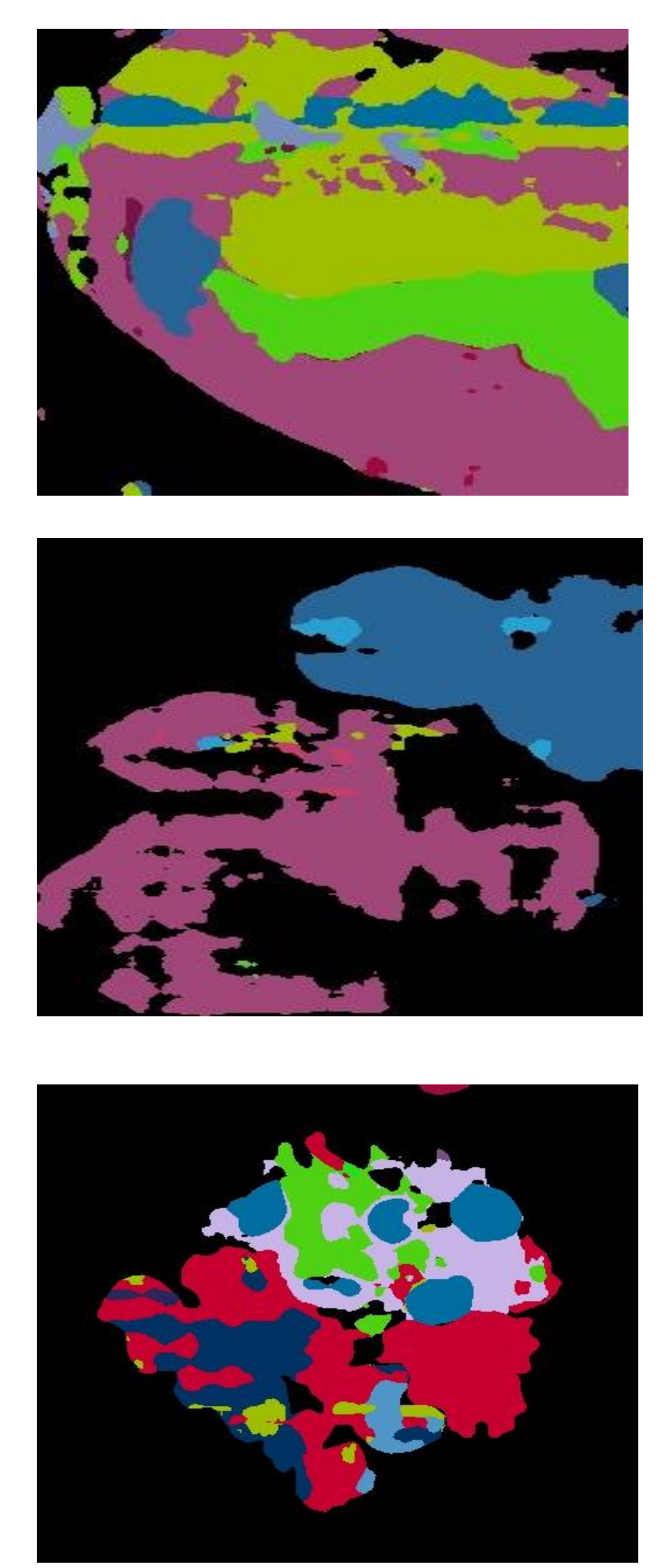}}
	\subfigure[Labels-Masks] {
		\includegraphics[width=0.18\textwidth]{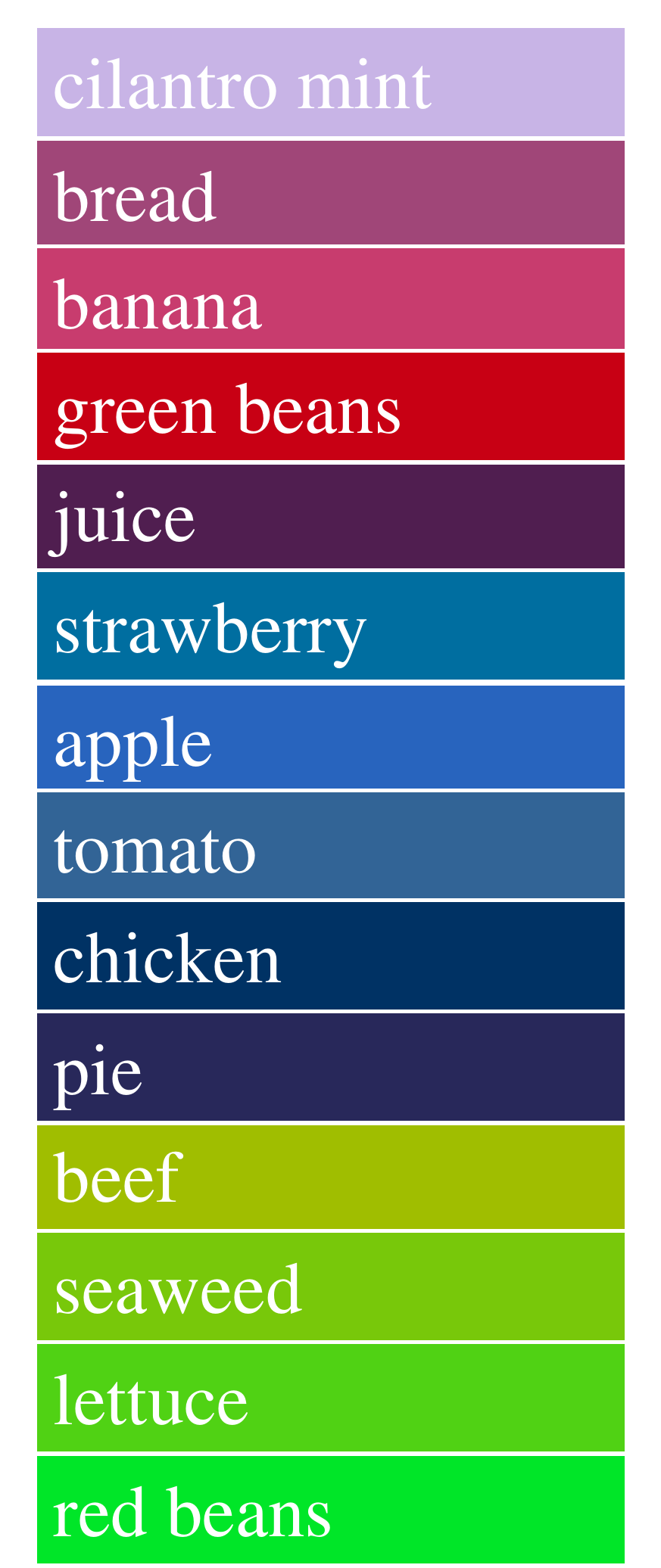}}
	\caption{Visualization results on FoodSeg103. ReLeM-CCNet can make more accurate predictions.} 	\label{fig:vis}
\end{figure*}

\vspace{-2mm}
\subsection{Results and Observations}
The experiment results of CCNet, FPN and SeTR on FoodSeg103 are shown in Table \ref{tab:ab}.

\begin{table}[tph]
		\centering
		\resizebox{0.45\textwidth}{!}{
		\begin{tabular}{|l|c|c|c|c|}
			\hline \hline
			Methods\;\; & mIoU &  mAcc  &Model Size\\
			\hline
			CCNet~\cite{huang2018ccnet} (ResNet-50)  &35.5 & 45.3 &381M\\
			ReLeM-CCNet (LSTM) &\textbf{36.8} &  47.4 &381M\\
			ReLeM-CCNet (Transformer) & 36.0 & 46.5 &381M\\
			\hline
			FPN~\cite{kirillov2019panoptic} (ResNet-50) &27.8 & 38.2 &218M\\
			ReLeM-FPN (LSTM)  &\textbf{29.1} &39.8 &218M\\
			ReLeM-FPN (Transformer)  &  28.9& 39.7 &218M\\
			\hline
			SeTR~\cite{SETR}, (ViT-16/B) &  41.3 & 52.7 &723M\\
			ReLeM-SeTR (LSTM)  & \textbf{43.9}&57.0&723M\\
			ReLeM-SeTR (Transformer) &43.2  &55.7  &723M\\
			\hline
			\hline
		\end{tabular}}
		\caption{Semantic segmentation results of our ReLeM plugged into three baseline methods (on the FoodSeg103 dataset). We implement two variants of ReLeM using LSTM and Transformer, respectively, to encode recipes. 
		}\label{tab:ab}
		\vspace{-4mm}
	\end{table}
\vspace{-1mm}
The Segmenters of all CCNet, FPN and SeTR achieve significant improvements when incorporating with either LSTM-based or transformer-based ReLeM (1.3\%, 1.3\% and 2.6\% improvement). 
This confirms that ReLeM is effective in enhancing both convolution based and transformer based semantic segmentation models.
Besides, we can see that the performance of using LSTM-based ReLeM is consistently superior than using transformer-based ReLeM across all the model configurations. 
\subsection{Comparing FoodSeg103 with Cityscapes} 
We compare the food image segmentation task with conventional semantic segmentation to compare the degree of difficulty of the two types of segmentation tasks.
We include three types of state-of-the-art segmentation algorithms, CCNet, SeTR and FPN.  They are evaluated on FoodSeg103 and Cityscapes~\cite{Cordts2016Cityscapes} datasets. Cityscapes contains around 5,000 images captured on the streets of German cities, and 20 types of objects as segmentation targets.
As we can see from Table~\ref{tab:cityscape}, all baseline methods achieve satisfactory results on Cityscapes, but suffer significant performance drops on our FoodSeg103. This indirectly shows the greater level of difficulty in the food image segmentation problem.

\begin{table}[thp]
		\centering
		\resizebox{0.43\textwidth}{!}
		{
		\begin{tabular}{|l|c|c|c|}
			\hline 
			Methods\;\; & Cityscapes & FoodSeg103 & \emph{gap}\\
			\hline
			CCNet &79.0 & 35.0 & 34.0 \\
			Sem-FPN &74.5 &27.8 & 46.7 \\
			SeTR &77.9 &41.3 &36.6 \\
			\hline
			\hline
		\end{tabular}}
		\caption{
		Semantic segmentation results on Cityscape~\cite{Cordts2016Cityscapes} and our FoodSeg103, showing that our FoodSeg103 is much more challenging than the object image dataset for the task of semantic segmentation.
		}
		\vspace{-5mm}
		\label{tab:cityscape}
	\end{table}

\begin{table}[thp]
		\centering
		\resizebox{0.45\textwidth}{!}
		{
		\begin{tabular}{|l|c|c|c|}
			\hline 
			Methods\;\;  &mIoU & mAcc & aAcc \\
			\hline
			CCNet &  28.6 & 47.8 & 78.9\\
			ReLeM-CCNet & 29.2 & 47.5 &79.3\\
			CCNet-Finetune &41.3 & 53.8 & 87.7\\
			ReLeM-CCNet-Finetune &47.1 &59.5&85.5\\
			\hline
			FPN & 21.9 & 41.7 & 75.5\\
			ReLeM-FPN & 22.9 &42.3 & 77.0\\
			FPN-Finetune & 27.1&38.0&82.6\\
			ReLeM-FPN-Finetune &30.8 &40.7 & 78.9\\
			\hline
			\hline
		\end{tabular}}
		\caption{Cross-domain adaptation results. We use LSTM based ReLeM.}
		\vspace{-5mm}
		\label{tab:adapt}
	\end{table}

\vspace{-3mm}
\subsection{Qualitative Examples}
In Figure \ref{fig:vis}, we show some qualitative results of using CCNet and ReLeM-CCNet on the testing set of FoodSed103. 
The first two rows clearly show that ReLeM-CCNet produces more accurate and detailed predictions than the vanilla CCNet, demonstrating the effectiveness of ReLeM. 
In the last row, we show a failure case.
It is actually a hard example with no clear boundaries among different ingredients.

\subsection{Cross-Domain Evaluation}
We conduct an out-domain model evaluation using the Asian food data set in FoodSeg154. 
With the model trained on FoodSeg103, we adapt it to the subset of FoodSeg154, the Asian food data set. Specifically, the Asia food set is evenly divided into the training and testing splits. 
We fine-tune the trained model on the training set and then run the model on the testing data.
In Table \ref{tab:adapt}, 
we show the performances of three models trained with the following settings:
1) without ReLeM, 2) with ReLeM and 3) with ReLeM and fine-tuned on the training split of the Asian food set. 
For the first two settings, we only evaluate the 62 classes in Asian food set overlapped with FoodSeg103, and for the last setting, we evaluate 112 classes (all).
From the results in Table~\ref{tab:adapt}, we observe that using ReLeM consistently outperforms baselines in both cases---with and without model fine-tuning on the training split of Asian food data.

\section{Conclusions}\label{sec:cl}
We construct a large-scale image dataset FoodSeg103 (and its extension FoodSeg154) for food image segmentation research.
We use around 10k images and annotate 60k segmentation masks in total, covering highly diverse appearances among 154 ingredients.
In addition, we propose a multi-modality based pre-training method ReLeM, and validate its effectiveness by incorporating three baseline semantic segmentation methods and conducting extensive experiments on the FoodSeg103, i.e., using the typical setting, as well as on the FoodSeg154, i.e., using the challenging cross-domain setting. 

\section{Acknowledgement}
This research is supported by the National Research Foundation, Singapore under its International Research Centres in Singapore Funding Initiative. Any opinions, findings and conclusions or recommendations expressed in this material are those of the author(s) and do not reflect the views of National Research Foundation, Singapore.
It is also partially supported by A*STAR under its AME YIRG Grant (Project No. A20E6c0101).

\bibliographystyle{ACM-Reference-Format}
\bibliography{foodseg}

\section*{Appendix: More Details of FoodSeg103 and FoodSeg154}

\subsection{Statistics}
\noindent{\textbf{Image Collection.}} For FoodSeg103, we first shuffle all the images and randomly select 70\% images (4983 images) as training set and the left 30\% images as testing set. For Asian Set, we randomly sample 50\% images (1186 images) for each dish class, and the left 50\% are used for testing. The basic information of training and testing set is listed in Table \ref{tab:overview_stat}, and the more detailed statistic can be found in Table \ref{tab:detailed_stat}. In our experiments, we use FoodSeg103 for in-domain training and testing, and use the additional Asian set for out-domain evaluation.

\begin{figure*}[htp]
	\centering
	\includegraphics[width=1\textwidth]{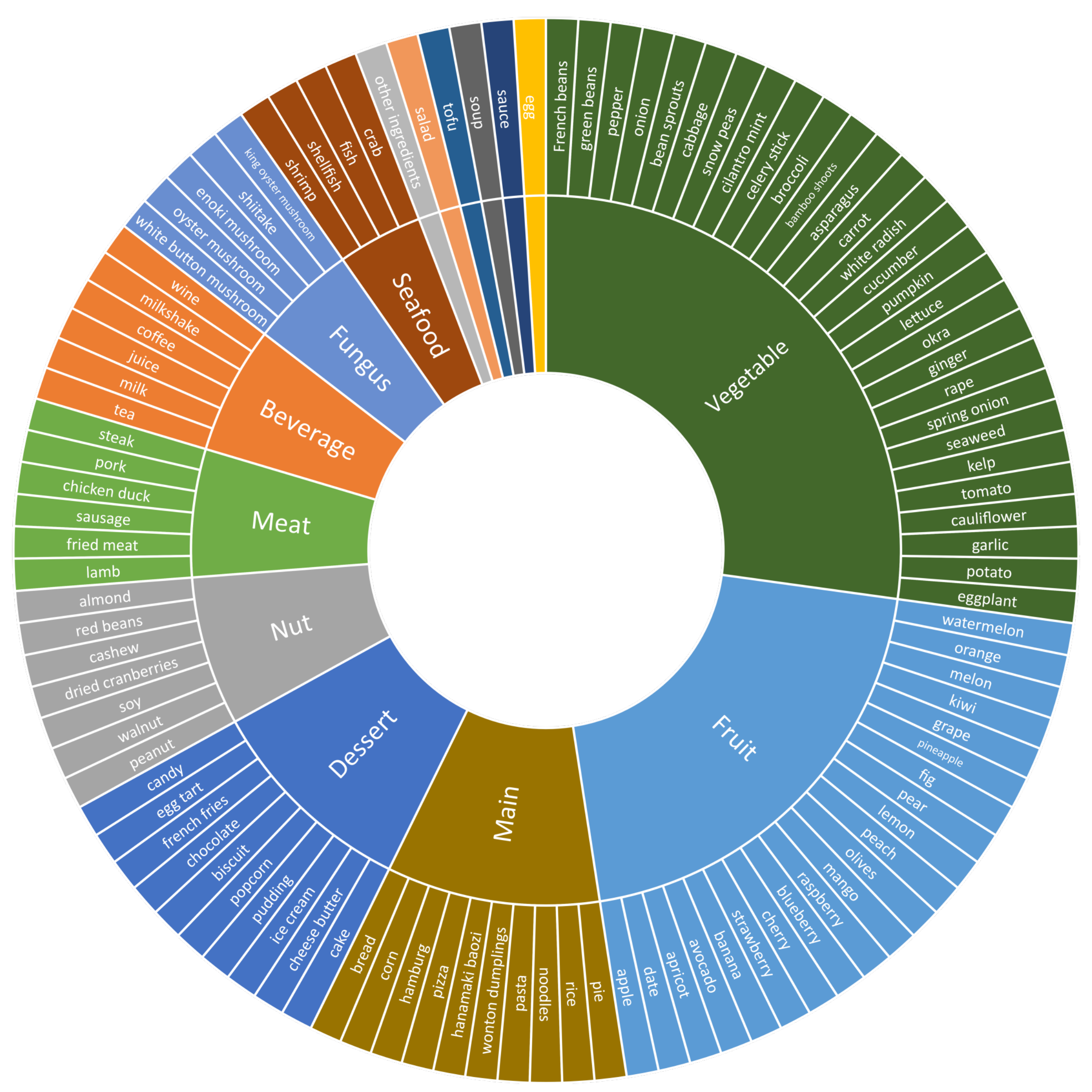}
	\caption{The dataset structure of FoodSeg103. The inner circle plots the super classes and the outer circle plots the corresponding sub-classes.} 	\label{fig:ont}
\end{figure*}

\noindent{\textbf{Structure of FoodSeg103}} FoodSeg103 contains 103 ingredient categories which belong to 15 super categories. In Figure \ref{fig:ont}, we show the dataset structure of FoodSeg103, where the inner circle plots the names of super classes, and the outer circle plots the corresponding ingredient categories. 

\subsection{Visualization}
\begin{figure*}[htp]
	\centering
	\includegraphics[width=1\textwidth]{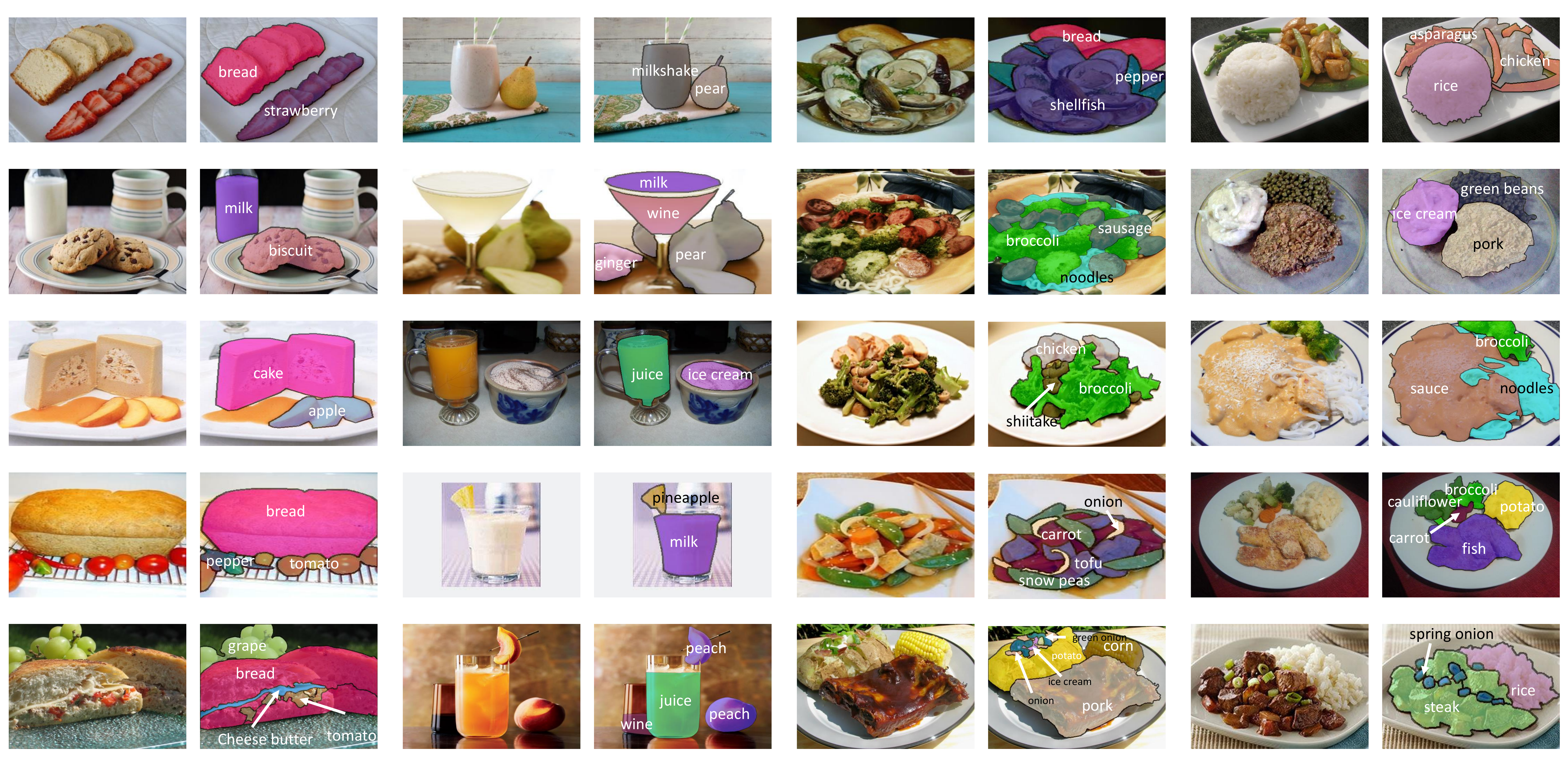}
	\caption{More annotation examples of FoodSeg103. The source images are in the left hand, while the annotation masks are in the right hand.} 	\label{fig:overview}
\end{figure*}

\noindent{\textbf{Visualization of FoodSeg103.}} In Figure \ref{fig:overview}, we show more visualization examples of the source image and its corresponding mask annotation in FoodSeg103.

\subsection{Analysis on Transformer-based Models}
Vision Transformers have been intensively studied recently, and a bunch of new algorithms have been proposed. The new proposed vision transformers have achieved significantly better performance than conventional CNN-based models in multiple vision tasks. 
In this section, we explore the performance of applying vision transformers into food image segmentation task. We adopt the vision transformers: ViT~\cite{dosovitskiy2021an}, Swin~\cite{liu2021Swin} and PVT~\cite{wang2021pyramid} as segmentation encoders. We follow the default design of decoders, where FPN is used in PVT models and UperNet~\cite{xiao2018unified} is used in Swin models. For ViT models, we use the two default settings in SeTR: Naive and MLA, as decoders. All the models are trained with the default learning settings with 80k iterations.
 
The results are shown in Table \ref{tab:app_tr}. ReLeM-variants show consistent improvement on both PVT and ViT-Naive models (0.7\% and 2.6\% improvement). However, in ViT-MLA model, the baseline shows better performance. In MLA decoder, feature maps from different level transformer encoders are integrated for final prediction. In ReLeM, however, only the last feature map is extracted for recipe learning. We argue ReLeM can also learn strong multi-level representation by extracting feature maps of different levels for recipe learning, and we leave it as the future work. In addition, larger backbones cannot guarantee improvement and may even hurt the performances (44.5\% vs 45.1\% in ViT, and 41.2\% vs 41.6\% in Swin). Besides, Swin achieves much better performance than ViT in other vision tasks~\cite{liu2021Swin}, but in food image segmentation, the performance of Swin is much worse than ViT models, even with more parameters. These results show that food image segmentation task is more challenging and naively boosting the power of backbone cannot guarantee performance gain. Finally, decoders play important roles in transformer-based segmenters, but few efforts have been made to design a food-aware decoders, which is also an important research problem in the future.

\begin{table}[htbp]
  \centering
  \resizebox{0.48\textwidth}{!}
  {
    \begin{tabular}{lrrrrrrrr}
    \toprule
    \hline
    && \multicolumn{3}{c}{\# Images} && \multicolumn{3}{c}{\# Ingredients}  \\
    \cmidrule{3-5}
    \cmidrule{7-9}
    Datasets       && {Train} & {Test} & {Total} && {Train} & {Test} & {Total} \\
       \hline
    FoodSeg103 && 4,983  &  2,135 & 7,118   && 29,530&12,567  & 42,097 \\
    Asian Set && 1,186  &  1,186  & 2,372  && 8,795  & 8,881 & 17,676 \\
    FoodSet154 && 6,169  &  3,321 & 9,490 &&  38,325& 21,448  & 59,773 \\
    \hline
    \end{tabular}%
    }
  \caption{Statistic of training and testing set for FoodSeg103, Asian Set and FoodSeg154.}
  \label{tab:overview_stat}%
\end{table}%

\begin{table}[htp]
  \centering
  \resizebox{0.48\textwidth}{!}
	{
    \begin{tabular}{|l|c||l|c||l|c|}
    \hline
    S-classes & Number &S-classes & Number&S-classes & Number\\
    \hline
    Dessert & 3913   &Meat  & 4956   &Soy   & 148\\
    \hline
    Beverage & 844 & Condiment & 1543 & Vegetable & 15719\\
    \hline
    Nut   & 912 & Seafood & 920   &Fungus & 592\\
    \hline
    Egg   & 424  & Soup  & 121 & Salad & 23 \\
    \hline
    Fruit & 6007 & Main  & 5634 & Others & 341\\
    \hline
    \hline
    \end{tabular}
    }%
    \caption{The ingredient number of all super classes in FoodSeg103.}
  \label{tab:app_103_super}%
\end{table}%

\begin{table}[tph]
		\centering
		\resizebox{0.48\textwidth}{!}{
		\begin{tabular}{|l|c|c|c|c|c|}
			\hline \hline
			Encoder\;\; &Decoder& mIoU &  mAcc & Model Size \\
			\hline
			PVT-S & FPN & 31.3 & 43.0 & 202M\\
			ReLeM-PVT-S & FPN & 32.0 & 44.1 & 202M\\
			\hline
			ViT-16/B & Naive & 41.3 &52.7 & 723M\\
			ReLeM-ViT-16/B & Naive & 43.9 &57.0 &723M\\
			\hline
			ViT-16/B & MLA & 45.1 &57.4 &711M\\
			ReLeM-ViT-16/B & MLA & 43.3 &55.9 &711M\\
			ViT-16/L & MLA & 44.5 &56.6 &2.4G\\
			\hline
			Swin-S & Uper &41.6& 53.6 &931M\\
			Swin-B & Uper &41.2& 53.9 & 1.4G\\

			\hline
			\hline
		\end{tabular}}
		\caption{Semantic segmentation results of different vision transformers. All models are trained based on the default learning settings with 4 images per batch for 80k iterations. ``S'', ``B'' and ``L'' denote ``Small'', ``Base'' and ``Large'' models respectively.
		}\label{tab:app_tr}
	\end{table}

\begin{table*}[htbp]
  \centering
  \resizebox{1.0\textwidth}{!}
	{
    \begin{tabular}{rcccccccccccc}
    \toprule
    \hline
    &&&\multicolumn{3}{c}{FoodSeg103}&&\multicolumn{2}{c}{Asian Set} &&\multicolumn{3}{c}{FoodSeg154}\\
    \cmidrule{4-6}
    \cmidrule{8-9}
    \cmidrule{11-13}
    {Class Id} & {{Class Name}} && {Train} & {Test} & Total && Train & Test& &\multicolumn{3}{c}{Total}    \\
    \hline
    1     & {candy} && 58    & 43    & 101    && 0     & 0  &&  \multicolumn{3}{c}{101} \\
   
    2     & {egg tart} && 8     & 6     & 14     && 0     & 0    & & \multicolumn{3}{c}{14} \\
   
    3     & {french fries} && 190   & 87    & 277   & & 95    & 83   &  & \multicolumn{3}{c}{455} \\
  
    4     & {chocolate} && 158   & 59    & 217   & & 0     & 0     & & \multicolumn{3}{c}{217} \\
    
    5     & {biscuit} && 393   & 122   & 515   & & 4     & 1     && \multicolumn{3}{c}{520} \\
    
    6     & {popcorn} && 37    & 11    & 48    & & 0     & 0     && \multicolumn{3}{c}{48} \\
   
    7     & {pudding} && 5     & 1     & 6     & & 0     & 0     && \multicolumn{3}{c}{6} \\
   
    8     & {ice cream} && 927   & 401   & 1328  & & 48    & 50    && \multicolumn{3}{c}{1426} \\
    
    9     & {cheese butter} && 461   & 198   & 659  &  & 19    & 14   & & \multicolumn{3}{c}{692} \\
   
    10    & {cake} & & 535  & 213   & 748   & & 0     & 0     && \multicolumn{3}{c}{748} \\
   
    11    & {wine} & & 117   & 50    & 167    && 15    & 19    && \multicolumn{3}{c}{201} \\
   
    12    & {milkshake} && 107   & 32    & 139   & & 0     & 0     && \multicolumn{3}{c}{139} \\
  
    13    & {coffee} && 136   & 62    & 198    && 8     & 12    && \multicolumn{3}{c}{218} \\
    
    14    & {juice} && 157   & 64    & 221    && 71    & 72    && \multicolumn{3}{c}{364} \\
    
    15    & {milk} && 48    & 36    & 84    & & 5     & 4     && \multicolumn{3}{c}{93} \\
    
    16    & {tea} && 29    & 6     & 35     && 15    & 6     && \multicolumn{3}{c}{56} \\
   
    17    & {almond} && 268   & 74    & 342   & & 0     & 0    & & \multicolumn{3}{c}{342} \\
    
    18    & {red beans} && 46    & 27    & 73     && 0     & 0     && \multicolumn{3}{c}{73} \\
    
    19    & {cashew} && 44    & 43    & 87    & & 0     & 0     && \multicolumn{3}{c}{87} \\
    
    20    & {dried cranberries} & & 79    & 55    & 134    && 0     & 0     && \multicolumn{3}{c}{134} \\
   
    21    & {soy} && 41    & 18    & 59    & & 0     & 0     && \multicolumn{3}{c}{59} \\
   
    22    & {walnut} && 100   & 81    & 181   & & 0     & 0     && \multicolumn{3}{c}{181} \\
  
    23    & {peanut} && 16    & 20    & 36    & & 93    & 95    && \multicolumn{3}{c}{224} \\
  
    24    & {egg} && 321   & 103   & 424   & & 162   & 161   && \multicolumn{3}{c}{747} \\
   
    25    & {apple} && 195   & 80    & 275   & & 29    & 49    && \multicolumn{3}{c}{353} \\
    
    26    & {date} && 14    & 3     & 17     && 51    & 43    && \multicolumn{3}{c}{111} \\
  
    27    & {apricot} && 39    & 18    & 57    & & 0     & 0     && \multicolumn{3}{c}{57} \\
   
    28    & {avocado} && 104   & 35    & 139   & & 9     & 19    && \multicolumn{3}{c}{167} \\
  
    29    & {banana} && 160   & 101   & 261   & & 0     & 0     && \multicolumn{3}{c}{261} \\
    
    30    & {strawberry} && 745   & 391   & 1136  & & 3     & 4     && \multicolumn{3}{c}{1143} \\
    
    31    & {cherry} && 474   & 140   & 614   & & 0     & 0     && \multicolumn{3}{c}{614} \\
    
    32    & {blueberry} && 559   & 218   & 777   & & 0     & 0     && \multicolumn{3}{c}{777} \\
    
    33    & {raspberry} && 108   & 59    & 167  &  & 0     & 0     && \multicolumn{3}{c}{167} \\
    
    34    & {mango} && 80    & 25    & 105    && 0     & 0     && \multicolumn{3}{c}{105} \\
    
    35    & {olives} && 98    & 44    & 142    && 0     & 0     & &\multicolumn{3}{c}{142} \\
    
    36    & {peach} && 137   & 29    & 166    && 0     & 0     && \multicolumn{3}{c}{166} \\
   
    37    & {lemon} && 609   & 263   & 872   & & 106   & 99    && \multicolumn{3}{c}{1077} \\

    \hline
    
    \end{tabular}
    }%
\end{table*}%

\begin{table*}[htbp]
  \centering
    \resizebox{1.0\textwidth}{!}
	{
    \begin{tabular}{rcccccccccccc}
    \toprule
    \hline
    &&&\multicolumn{3}{c}{FoodSeg103}&&\multicolumn{2}{c}{Asian Set} &&\multicolumn{3}{c}{FoodSeg154}\\
    \cmidrule{4-6}
    \cmidrule{8-9}
    \cmidrule{11-13}
    {Class Id} & {{Class Name}} && {Train} & {Test} & Total && Train & Test& &\multicolumn{3}{c}{Total}    \\
    \hline
        
    38    & {pear} && 55    & 21    & 76    & & 0     & 0     & &\multicolumn{3}{c}{76} \\
    
    39    & {fig} && 51    & 9     & 60    & & 0     & 0     &&\multicolumn{3}{c}{60}  \\
    
    40    & {pineapple} && 205   & 81    & 286    && 32    & 37    && \multicolumn{3}{c}{355} \\
    41    & {grape} && 189   & 48    & 237  & & 0     & 0     &&\multicolumn{3}{c}{237} \\
    
    42    & {kiwi} && 69    & 21    & 90    && 0     & 0     && \multicolumn{3}{c}{90} \\
    
    43    & {melon} && 44    & 7     & 51    && 0     & 0     && \multicolumn{3}{c}{51} \\
   
    44    & {orange} && 283   & 110   & 393  & & 54    & 48    && \multicolumn{3}{c}{495} \\
    
    45    & {watermelon} && 68    & 18    & 86  &  & 0     & 0     &&\multicolumn{3}{c}{86} \\
   
    46    & {steak} && 987   & 483   & 1470 & & 0     & 0     &&\multicolumn{3}{c}{1470} \\
  
    47    & {pork} && 646   & 261   & 907  & & 0     & 0     & &\multicolumn{3}{c}{907} \\
   
    48    & {chicken duck} && 1160  & 508   & 1668  && 0     & 0    & & \multicolumn{3}{c}{1668} \\
   
    49    & {sausage} && 372   & 93    & 465  & & 32    & 34    && \multicolumn{3}{c}{531} \\
    
    50    & {fried meat} && 209   & 118   & 327   && 0     & 0     && \multicolumn{3}{c}{327} \\
   
     51    & {lamb} && 85    & 34    & 119  & & 0     & 0     && \multicolumn{3}{c}{119} \\
   
    52    & {sauce} && 1124  & 419   & 1543  && 19    & 15    && \multicolumn{3}{c}{1577} \\
    
    53    & {crab} && 19    & 11    & 30    && 38    & 37    && \multicolumn{3}{c}{105} \\
   
    54    & {fish} && 348   & 138   & 486  & & 103   & 126   && \multicolumn{3}{c}{715} \\
    
    55    & {shellfish} && 77    & 27    & 104  & & 37    & 40    && \multicolumn{3}{c}{181} \\
    
    56    & {shrimp} && 211   & 89    & 300   && 51    & 54    && \multicolumn{3}{c}{405} \\
   
    57    & {soup} && 92    & 29    & 121   && 0     & 0     && \multicolumn{3}{c}{121} \\
    
    58    & {bread} && 1698  & 738   & 2436  && 49    & 40    && \multicolumn{3}{c}{2525} \\
   
    59    & {corn} && 411   & 170   & 581   && 29    & 35    && \multicolumn{3}{c}{645} \\
    
    60    & {hamburg} && 7     & 1     & 8    & & 0     & 0     && \multicolumn{3}{c}{8} \\
   
    61    & {pizza} && 83    & 22    & 105   && 0     & 0     && \multicolumn{3}{c}{105} \\
    
    62    & { hanamaki baozi} && 22    & 14    & 36   & & 0     & 0     && \multicolumn{3}{c}{36} \\
   
    63    & {wonton dumplings} && 10    & 10    & 20    && 165   & 149   && \multicolumn{3}{c}{334} \\
    
    64    & {pasta} && 171   & 59    & 230   && 18    & 3     && \multicolumn{3}{c}{251} \\
    
    65    & {noodles} && 337   & 140   & 477   && 811   & 836   && \multicolumn{3}{c}{2124} \\
    
    66    & {rice} && 655   & 277   & 932  & & 294   & 306   && \multicolumn{3}{c}{1532} \\
    
    67    & {pie} && 563   & 246   & 809   && 20    & 17    && \multicolumn{3}{c}{846} \\
    
    68    & {tofu} && 111   & 37    & 148   && 73    & 57    && \multicolumn{3}{c}{278} \\
    
    69    & {eggplant} & &34    & 9     & 43   & & 38    & 12    && \multicolumn{3}{c}{93} \\
   
    70    & {potato} && 1041  & 400   & 1441  && 110   & 111   && \multicolumn{3}{c}{1662} \\
   
    71    & {garlic} && 143   & 29    & 172   && 40    & 36    && \multicolumn{3}{c}{248} \\
   
    72    & {cauliflower} && 237   & 100   & 337   && 43    & 32    && \multicolumn{3}{c}{412} \\
   
    73    & {tomato} && 1404  & 687   & 2091  && 124   & 100   && \multicolumn{3}{c}{2315} \\
   
    74    & {kelp} && 4     & 5     & 9     && 0     & 0     && \multicolumn{3}{c}{9} \\
   
    \hline
    
    \end{tabular}%
    }
  \label{tab:addlabel}%
\end{table*}%

\begin{table*}[htbp]
  \centering
  \resizebox{1.0\textwidth}{!}
	{
    \begin{tabular}{rcccccccccccc}
    \toprule
    \hline
    &&&\multicolumn{3}{c}{FoodSeg103}&&\multicolumn{2}{c}{Asian Set} &&\multicolumn{3}{c}{FoodSeg154}\\
    \cmidrule{4-6}
    \cmidrule{8-9}
    \cmidrule{11-13}
    {Class Id} & {{Class Name}} && {Train} & {Test} & Total && Train & Test& &\multicolumn{3}{c}{Total}    \\
    \hline
     75    & {seaweed} && 16    & 10    & 26    && 29    & 29    && \multicolumn{3}{c}{84} \\
    
    76    & {spring onion} && 285   & 113   & 398  & & 556   & 561   && \multicolumn{3}{c}{1515} \\
   
    77    & {rape} && 59    & 23    & 82    && 360   & 429   && \multicolumn{3}{c}{871} \\
   
    78    & {ginger} && 25    & 12    & 37    && 24    & 34    && \multicolumn{3}{c}{95} \\
    
    79    & {okra} && 35    & 9     & 44   & & 31    & 18    && \multicolumn{3}{c}{93} \\
    
    80    & {lettuce} && 748   & 338   & 1086  && 245   & 230   && \multicolumn{3}{c}{1561} \\
    81    & {pumpkin} && 114   & 25    & 139   && 0     & 0     && \multicolumn{3}{c}{139} \\
    
    82    & {cucumber} && 568   & 267   & 835   && 234   & 203  & & \multicolumn{3}{c}{1272} \\
   
    83    & {white radish} && 56    & 34    & 90    && 63    & 52    && \multicolumn{3}{c}{205} \\
    
    84    & {carrot} && 1407  & 670   & 2077 & & 156   & 143   && \multicolumn{3}{c}{2376} \\
    
    85    & {asparagus} && 325   & 139   & 464   && 24    & 23    && \multicolumn{3}{c}{511} \\
    
    86    & {bamboo shoots} && 8     & 7     & 15    && 0     & 0    & & \multicolumn{3}{c}{15} \\
    
    87    & {broccoli} && 966   & 427   & 1393  && 35    & 49   & & \multicolumn{3}{c}{1477} \\
    
    88    & {celery stick} && 233   & 91    & 324  & & 36    & 35   & & \multicolumn{3}{c}{395} \\
    
    89    & {cilantro mint} && 1045  & 466   & 1511  && 323   & 320 &  & \multicolumn{3}{c}{2154} \\
    
    90    & {snow peas} && 103   & 49    & 152   && 6     & 16    && \multicolumn{3}{c}{174} \\
    
    91    & { cabbage} && 139   & 39    & 178   && 25    & 13    && \multicolumn{3}{c}{216} \\
    
    92    & {bean sprouts} && 35    & 20    & 55    && 34    & 34  &  & \multicolumn{3}{c}{123} \\
    
    93    & {onion} && 732   & 304   & 1036 & & 85    & 103  & & \multicolumn{3}{c}{1224} \\
    
    94    & {pepper} && 552   & 242   & 794   && 189   & 191  && \multicolumn{3}{c}{1174} \\
    
    95    & {green beans} && 237   & 125   & 362   && 40    & 37   & & \multicolumn{3}{c}{439} \\
    
    96    & {French beans} && 360   & 168   & 528   && 39    & 34   & & \multicolumn{3}{c}{601} \\
    
    97    & {king oyster mushroom} && 12    & 3     & 15   & & 0     & 0  &   & \multicolumn{3}{c}{15} \\
   
    98    & {shiitake} &&185   & 106   & 291   && 167   & 205   && \multicolumn{3}{c}{663} \\
   
    99    & {enoki mushroom} && 9     & 5     & 14   & & 25    & 31   & & \multicolumn{3}{c}{70} \\
    
    100   & {oyster mushroom} & &11    & 4     & 15   & & 0     & 0    & &\multicolumn{3}{c}{15} \\
  
    101   & {white button mushroom} && 195   & 62    & 257   && 35    & 26    && \multicolumn{3}{c}{318} \\
    
     102   & {salad} && 12    & 11    & 23   & & 0     & 0     && \multicolumn{3}{c}{23} \\
   
    103   & {other ingredients} && 230   & 111   & 341  & & 667   & 738   && \multicolumn{3}{c}{1746} \\
    
    104   & {water} && 0     & 0     & 0     && 2     & 4     && \multicolumn{3}{c}{6} \\
   
    105   & {goji berry} && 0     & 0     & 0    & & 33    & 50   & &\multicolumn{3}{c}{83} \\
   
    106   & {ribs} && 0     & 0     & 0    & & 148   & 135  & & \multicolumn{3}{c}{283} \\
   
    107   & {tripe} && 0     & 0     & 0    & & 31    & 36    &&\multicolumn{3}{c}{67} \\
    
    108   & {meat slices} && 0     & 0     & 0    & & 135   & 170  & & \multicolumn{3}{c}{305} \\
    
    109   & {minced meat} && 0     & 0     & 0   &  & 95    & 69   & & \multicolumn{3}{c}{164} \\
    
    110   & {pork belly} && 0     & 0     & 0   &  & 87    & 76   & & \multicolumn{3}{c}{163} \\
   
    111   & {pork intestine} && 0     & 0     & 0     && 16    & 16  &  & \multicolumn{3}{c}{32} \\
    
    112   & {pork skin} && 0     & 0     & 0     && 33    & 15  &  & \multicolumn{3}{c}{48} \\
    
    113   & {blood} && 0     & 0     & 0     && 4     & 4    & & \multicolumn{3}{c}{8} \\
   
    114   & {pork liver} && 0     & 0     & 0     && 26    & 16   & & \multicolumn{3}{c}{42} \\
    
    115   & shredded pork && 0     & 0     & 0    & & 25    & 34  &  &\multicolumn{3}{c}{59} \\
    
    \hline
    \end{tabular}%
    }
  \label{tab:addlabel}%
\end{table*}%

\begin{table*}[htbp]
  \centering
  \resizebox{1.0\textwidth}{!}
  {
   \begin{tabular}{rcccccccccccc}
    \toprule
    \hline
    &&&\multicolumn{3}{c}{FoodSeg103}&&\multicolumn{2}{c}{Asian Set} &&\multicolumn{3}{c}{FoodSeg154}\\
    \cmidrule{4-6}
    \cmidrule{8-9}
    \cmidrule{11-13}
    {Class Id} & {{Class Name}} && {Train} & {Test} & Total && Train & Test& &\multicolumn{3}{c}{Total}    \\
    \hline
     116   & chicken legs/duck legs && 0     & 0     & 0    & & 65    & 62   & & \multicolumn{3}{c}{127} \\
    
    117   & meat skewers& & 0     & 0     & 0     && 48    & 51   & & \multicolumn{3}{c}{99} \\
    
    118   & chicken feet && 0     & 0     & 0     && 32    & 33   & & \multicolumn{3}{c}{65} \\
   
    119   & barbecued pork && 0     & 0     & 0     && 125   & 102 &  & \multicolumn{3}{c}{227} \\
    
    120   & beef ball && 0     & 0     & 0     && 81    & 60    && \multicolumn{3}{c}{141} \\
    121   & poultry meat && 0     & 0     & 0    & & 235   & 234   && \multicolumn{3}{c}{469} \\
    
    122   & barbecued pork sauce && 0     & 0     & 0     && 69    & 73   & & \multicolumn{3}{c}{142} \\
    
    123   & caviar& & 0     & 0     & 0     && 24    & 22    && \multicolumn{3}{c}{46} \\
    
    124   & curry sauce && 0     & 0     & 0    & & 0     & 11    && \multicolumn{3}{c}{11} \\
   
    125   & satay sauce && 0     & 0     & 0    & & 36    & 45    && \multicolumn{3}{c}{81} \\
    
    126   & chili sauce && 0     & 0     & 0    & & 99    & 95    && \multicolumn{3}{c}{194} \\
   
    127   & ketchup && 0     & 0     & 0    & & 35    & 21    && \multicolumn{3}{c}{56} \\
   
    128   & salad sauce && 0     & 0     & 0    & & 16    & 20    && \multicolumn{3}{c}{36} \\
    
    129   & basil sauce && 0     & 0     & 0   &  & 30    & 25    && \multicolumn{3}{c}{55} \\
    
    130   & garlic sauce && 0     & 0     & 0   &  & 8     & 8     && \multicolumn{3}{c}{16} \\
   
    131   & cuttlefish && 0     & 0     & 0   &  & 4     & 3     && \multicolumn{3}{c}{7} \\
    
    132   & squid && 0     & 0     & 0    & & 32    & 31    && \multicolumn{3}{c}{63} \\
   
    133   & fish cakes && 0     & 0     & 0     && 78    & 100  & & \multicolumn{3}{c}{178} \\
    
    134   & fish Ball && 0     & 0     & 0    & & 220   & 205   && \multicolumn{3}{c}{425} \\
    
    135   & fish tofu && 0     & 0     & 0    & & 27    & 26    && \multicolumn{3}{c}{53} \\
    
    136   & fried fish && 0     & 0     & 0     && 76    & 66    && \multicolumn{3}{c}{142} \\
   
    137   & small dried fish && 0     & 0     & 0    & & 73    & 71   & & \multicolumn{3}{c}{144} \\
    
    138   & yut yiao && 0     & 0     & 0    & & 46    & 56    && \multicolumn{3}{c}{102} \\
    
    139   & porridge & &0     & 0     & 0     && 36    & 55    && \multicolumn{3}{c}{91} \\
    
    140   & fried banana leaves && 0     & 0     & 0   &  & 23    & 32    && \multicolumn{3}{c}{55} \\
   
    141   & rice cake && 0     & 0     & 0    & & 16    & 14    && \multicolumn{3}{c}{30} \\
   
    142   & yuba  && 0     & 0     & 0     && 27    & 29    && \multicolumn{3}{c}{56} \\
    
    143   & fried tofu && 0     & 0     & 0  &   & 11    & 24    && \multicolumn{3}{c}{35} \\
    
    144   & beancurd puff && 0     & 0     & 0    & & 26    & 33    && \multicolumn{3}{c}{59} \\
    
    145   & preserved vegetable && 0     & 0     & 0    & & 7     & 17   & & \multicolumn{3}{c}{24} \\
    
    146   & salted vegetables && 0     & 0     & 0    & & 32    & 25    && \multicolumn{3}{c}{57} \\
    
    147   & pea seedlings && 0     & 0     & 0     && 13    & 15    && \multicolumn{3}{c}{28} \\
    
    148   & kai lan && 0     & 0     & 0     && 6     & 11    && \multicolumn{3}{c}{17} \\
    
    149   & lotus root && 0     & 0     & 0     && 26    & 26    && \multicolumn{3}{c}{52} \\
   
    150   & amaranth && 0     & 0     & 0    & & 23    & 16   & & \multicolumn{3}{c}{39} \\
    
    151   & millet spicy && 0     & 0     & 0    & & 64    & 65   & & \multicolumn{3}{c}{129} \\
    
    152   & bitter gourd && 0     & 0     & 0    & & 16    & 17    && \multicolumn{3}{c}{33} \\
   
    153   & daylily & &0     & 0     & 0    & & 1     & 5     && \multicolumn{3}{c}{6} \\
    
    154   & agaric && 0     & 0     & 0     && 33    & 42    && \multicolumn{3}{c}{75} \\
    \hline
    -   & \textbf{Summary} && \textbf{29530}  & \textbf{12567}     & \textbf{42097}     &&  \textbf{8795}   &    \textbf{8881} &&\multicolumn{3}{c}{\textbf{59773}} \\
        \hline
    \end{tabular}%
    }
\caption{Statistic of ingredients per class for FoodSeg103, Asian set and FoodSeg154. }
  \label{tab:detailed_stat}%
\end{table*}%

\end{document}